\documentclass[10pt,twocolumn,letterpaper]{article}

\usepackage[pagenumbers]{cvpr} % To force page numbers, e.g. for an arXiv version

\usepackage[dvipsnames]{xcolor}

\definecolor{cvprblue}{rgb}{0.21,0.49,0.74}
\usepackage[pagebackref,breaklinks,colorlinks,citecolor=cvprblue]{hyperref}

\usepackage{multirow}
\usepackage{colortbl}
\usepackage{caption}[font={small}]
\usepackage{tcolorbox}
\usepackage{xcolor}[html]

\def\paperID{682} % *** Enter the Paper ID here
\def\confName{CVPR}
\def\confYear{2024}
\usepackage{amsmath}

\newcommand{\name}{Digital Life Project\xspace}
\newcommand{\short}{DLP\xspace}
\newcommand{\brain}{SocioMind\xspace}
\newcommand{\body}{MoMat-MoGen\xspace}
\newcommand{\plot}{episode\xspace}
\newcommand{\Fig}{Fig.\xspace}

\newcommand{\Sec}{Sec.\xspace}
\newcommand{\Supp}{Supplementary Material\xspace}

\title{\name: Autonomous 3D Characters with Social Intelligence}

\author{
Zhongang Cai$^{*,1,2,3}$, Jianping Jiang$^{*,2}$, Zhongfei Qing$^{*,2}$, Xinying Guo$^{*,1}$, Mingyuan Zhang$^{*,1}$, \\
Zhengyu Lin$^{2}$, Haiyi Mei$^{2}$, Chen Wei$^{2}$, Ruisi Wang$^{1,2}$, Wanqi Yin$^{2}$, Xiangyu Fan$^{2}$, Han Du$^{2}$, \\
Liang Pan$^{1,3}$, Peng Gao$^{2}$, Zhitao Yang$^{2}$, Yang Gao$^{2}$, Jiaqi Li$^{2}$, Tianxiang Ren$^{2}$, Yukun Wei$^{2}$, \\
Xiaogang Wang$^{2}$, Chen Change Loy$^{1}$, Lei Yang$^{\dagger,2,3}$, Ziwei Liu$^{\dagger,1}$ \\
$^{1}$S-Lab, Nanyang Technological University, $^{2}$SenseTime Research, $^{3}$Shanghai AI Laboratory \\
{\small $^{*}$ Equal Contributions, $^{\dagger}$ Corresponding Author} \\
\url{https://digital-life-project.com}
}

\begin{document}

% \maketitle
\twocolumn[{
    \renewcommand\twocolumn[1][]{#1}
    \maketitle
    \vspace{-25pt}
    \begin{center}
        \includegraphics[width=1\textwidth]{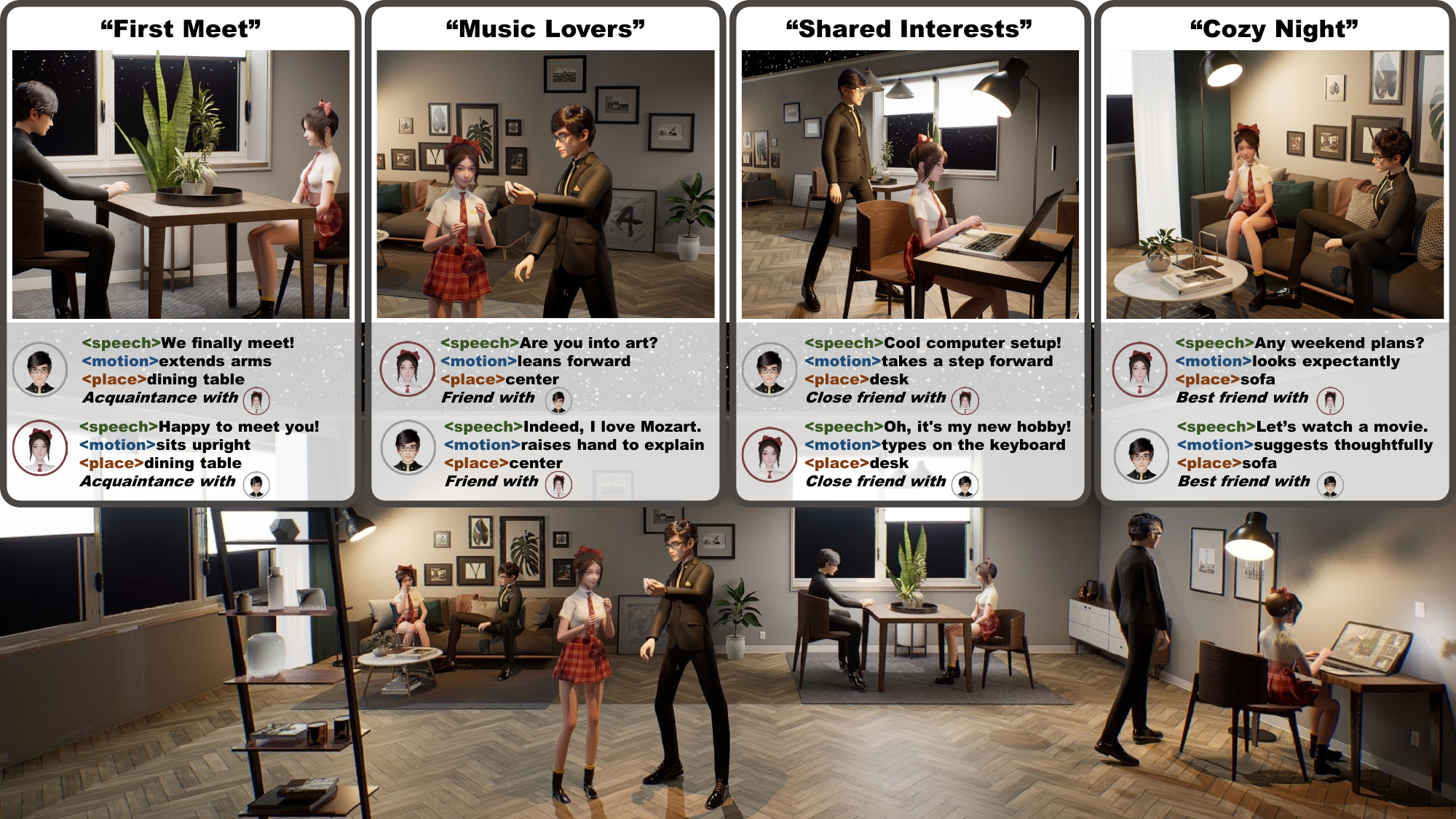}
        \vspace{-6mm}
        \captionof{figure}{\textbf{\name} empowers virtual characters to interact with each other using articulated body motions. We demonstrate the interaction of two characters across four occasions (\textit{episodes}) that leads to evolving relationship.}
        \label{fig:teaser}
    \end{center}
}]

\begin{abstract}
In this work, we present \textbf{\name}, a framework utilizing language as the universal medium to build autonomous 3D characters, who are capable of engaging in social interactions and expressing with articulated body motions, thereby simulating life in a digital environment. Our framework comprises two primary components:
\textbf{1) \brain:} a meticulously crafted \textit{digital brain} that models personalities with systematic few-shot exemplars, incorporates a reflection process based on psychology principles, and emulates autonomy by initiating dialogue topics; 
\textbf{2) \body:} a text-driven motion synthesis paradigm for controlling the character's \textit{digital body}. It integrates motion matching, a proven industry technique to ensure motion quality, with cutting-edge advancements in motion generation for diversity.
Extensive experiments demonstrate that each module achieves state-of-the-art performance in its respective domain. Collectively, they enable virtual characters to initiate and sustain dialogues autonomously, while evolving their socio-psychological states. Concurrently, these characters can perform contextually relevant bodily movements. Additionally, a motion captioning module further allows the virtual character to recognize and appropriately respond to human players' actions.
\end{abstract}
\vspace{-2mm}
\section{Introduction}
\label{sec:introduction}
\vspace{-2mm}

% Context
Recent advancements in Large Language Models (LLMs)~\cite{chatgpt, touvron2023llama} have transformed the landscape of human-computer interaction, catalyzing the emergence of innovative applications across various domains. Remarkably, many once far-fetched fantasies have gradually become tangible realities. In this work, the term \textit{\name} (\short), as envisioned in the recent science fiction blockbuster \textit{The Wandering Earth II}, is adopted to frame our endeavor. What qualifies as a digital life? From the psychological perspective, humans are composed of internal psychological processes (mind, such as thoughts) and external behaviors~\cite{james2007principles}. In this light, our objective is to harness the sophisticated capabilities of LLM to craft virtual 3D characters, that emulate the full spectrum of human psychological processes, and engage in diverse interactions with synthesized 3D body motions.

% Research gap
Recently, Park \etal introduced Generative Agents~\cite{Park2023GenerativeAgents} to advance AI agents capable of simulating human-like behavior. Despite the encouraging progress, this pioneering work is built upon many simplification of interaction: the agents are represented by pixelated 2D figures. Co-LLM-Agents~\cite{zhang2023building} aims to build collaborative embodied AI and includes 3D agents. However, the 3D agents are still constrained by a small set of actions and do not exhibit the capability to socialize. Existing works thus overlook the importance of sophisticated human body language, through which a crucial amount of information is conveyed~\cite{berger2010handbook, hall1966hidden, hall1973silent}. Moreover, there is a notable deficiency in the current modeling of social intelligence. This aspect is critical for the creation of characters that not only mimic human actions but also possess human-like thinking and emotional responses, even the ability to foster long-term relationships.

% DLP: brain, body, eyes
To achieve the aspirations of \short, we introduce a framework consisting of two essential components. 
\textbf{First}, the \brain which is a carefully designed ``digital brain", anchoring its design in rigorously applied psychological principles. Utilizing emergent abilities of LLMs~\cite{chatgpt, openai2023gpt4, wei2022emergent}, the brain generates high-level instructions and plans the character's behaviors. Notably, \brain introduces few-shot exemplars from psychological tests to form guiding instructions for personality modeling, utilizes social cognitive psychology theories in the memory reflection process, and designs a negotiation mechanism between characters for story progression.
\textbf{Second}, the ``digital body" that introduces the \body paradigm to address interactive motion synthesis, which exploits the complementary nature of motion matching~\cite{clavet2016motion} and motion generation~\cite{zhang2023remodiffuse}. Here, motion matching is a foundational technique in modern-day industry-level character animation that retrieves high-quality motion clips from a database to ensure motion quality, whereas motion generation is a line of works that rapidly gained popularity recently for their excellent ability to produce diverse human motions. 

% Experiment results and Applications
Experiment results demonstrate that \brain and \body outperform existing arts in their respective domains. Specifically, \brain demonstrates outstanding alignment between character behavior and psychological states (\eg., personality and relationship); \body is able to achieve a balance between motion quality and diversity. Equipped with both modules, we further show \short's controllability as manual editing of character attributes can result in semantically accurate and aesthetically realistic interactive motions. Moreover, we explore human-character interaction by developing a motion captioning module that translates monocular human video to motion description, thus enabling virtual characters to understand and appropriately respond to human players.

% The versatility of \short opens avenues to multiple potential applications. For example, \textbf{AI Filmmaking} where characters can bring a screenplay to life, transforming text narratives with interactive plots into actual production; In the \textbf{AI Society}, characters are granted full autonomy to behave on their own, to engage in dialogue and build a relationship with another character; 
% \textbf{Third}, eyes to translate human motions into a universal text representation, empowering the virtual characters to perceive and intelligently react to the gestures of human users.
% \textbf{AI Companion} features human users to immerse themselves in the virtual world, establishing a ``in-person" interaction with their digital counterpart. 

% Summary
In summary, we contribute \textbf{1)} \short, a framework to build autonomous 3D characters that possess social traits. It features \textbf{2)} \brain: a controllable psychology-based ``brain" to enable short-term interactive communication and long-term social evolution; \textbf{3)} \body: a ``body" that synthesizes high-quality and diverse interactive motions through synergizing motion matching and motion generation.
\section{Related Works}
\label{sec:related_works}

\subsection{Motion Synthesis}
Motion matching is widely employed in the industry to generate long-lasting, high-quality motion. The classic motion matching \cite{clavet2016motion} retrieves the segment that best matches the current pose and target trajectory. Learned motion matching \cite{holden2020learned} employs an auto-regressive neural network to predict the next motion state based on a given control signal. The Story-to-motion \cite{qing2023storytomotion} further incorporates semantic control through LLM and enhances transition using transformer models. Recently, significant strides have been made in motion generative models for text-driven motion generation. Early works aimed to establish a unified latent space for natural language and motion sequences~\cite{ahuja2019language2pose, ghosh2021synthesis, tevet2022motionclip, petrovich2022temos}. Guo et al.~\cite{guo2022generating}, TM2T~\cite{guo2022tm2t}, and T2M-GPT~\cite{zhang2023generating} employ an auto-regressive scheme to generate lengthy motion sequences. Diffusion-based generative models have demonstrated remarkable performance in leading benchmarks for the text-to-motion task. MotionDiffuse~\cite{zhang2022motiondiffuse}, MDM~\cite{tevet2022human}, and FLAME~\cite{kim2023flame} represent early attempts to apply the diffusion model to the text-driven motion generation field. Subsequent models such as MLD~\cite{chen2023executing}, ReMoDiffuse~\cite{zhang2023remodiffuse}, Fg-T2M~\cite{Wang_2023_ICCV}, and PhysDiff~\cite{Yuan_2023_ICCV} have further advanced this idea, achieving improved text-motion consistency, motion quality, and physical plausibility.

\begin{figure*}[t]
    \centering
    \vspace{-6mm}
    \includegraphics[width=\linewidth]{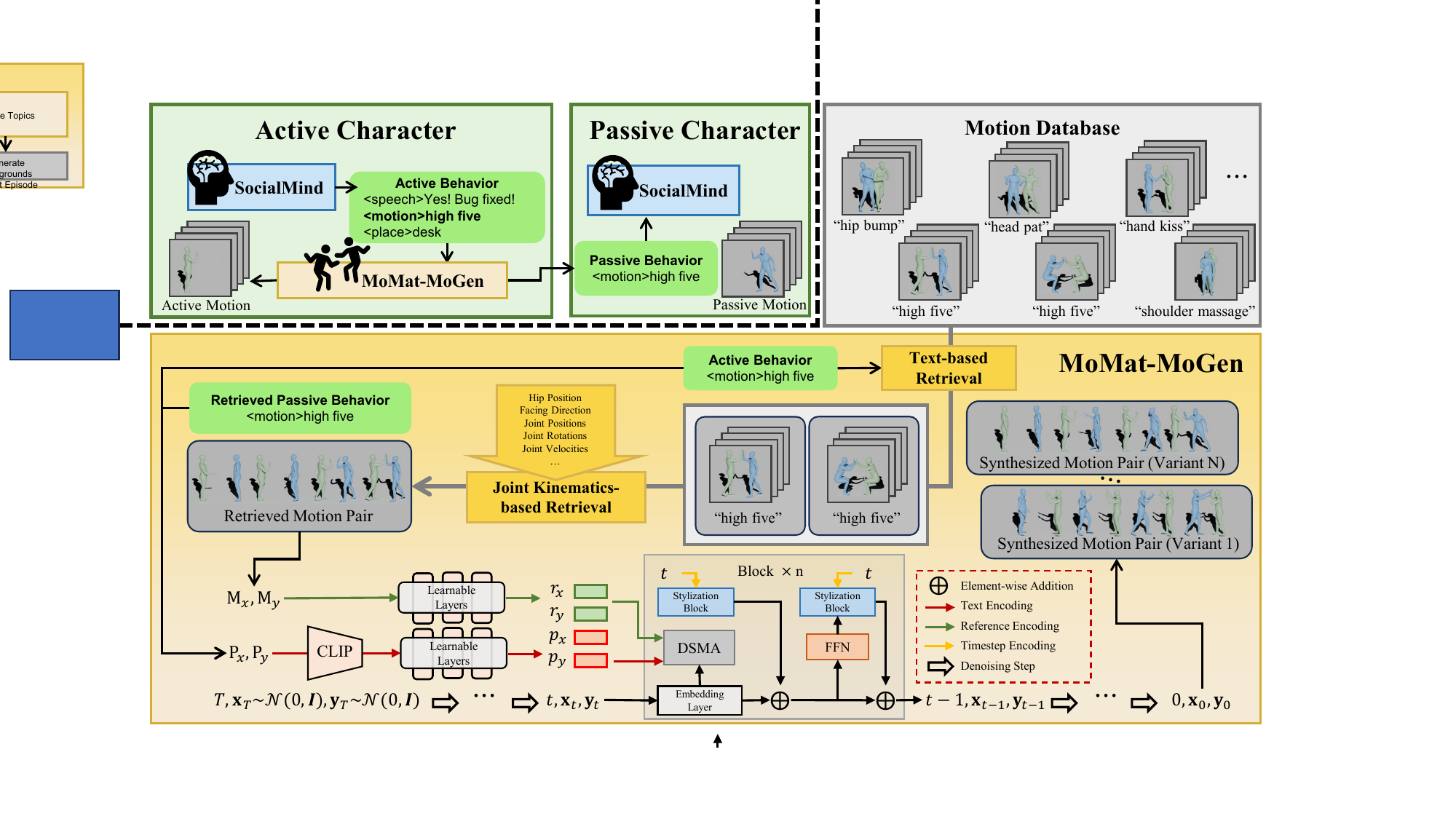}
    \vspace{-5mm}
    \caption{\textbf{\name} framework for interactive autonomous characters. The top left part depicts the Active-Passive Mechanism, and the rest of the figure illustrates \body. \brain is shown in details in \cref{fig:brain}.}
    \label{fig:method}
    \vspace{-2mm}
\end{figure*}

\subsection{LLM Agents}
\label{sec:llm-agent}
% [My brief idea: What distinguishes us from previous social agents are dimensions: interaction environment and purpose.]
With the emergent abilities of large language models (LLM) in reasoning, planning, and learning~\cite{du2021glm, touvron2023llama, openai2023gpt4, wei2022emergent}, LLMs swiftly evolve through three phases: the standalone primitive LLM, language agents~\cite{ahn2022can, huang2022inner} that directly interact with the environment via text, and cognitive language agents~\cite{wang2023voyager, park23generative, schick2023toolformer, yao2023react, yao2023tree} with internal cognitive structures~\cite{sumers2023cognitive}. 
% Leveraging the robust generalization abilities of LLM, LLM-agents have found widespread applications in social sciences, natural sciences, and computer engineering. 
Under the prime framework of cognitive language agents, the system design hinges on the intended application and objectives: reward systems for game agents~\cite{wang2023voyager, zhu2023ghost, xu2023werewolf}, chains of API calls for tool agents~\cite{schick2023toolformer, shen2023hugginggpt, patil2023gorilla}, and so forth.
% In the realm of building agents with social intelligence, information diffusion and social interaction depend on the space where the agent resides.
% to facilitate the formation of social relationships and information diffusion 
% However, grounding an LLM-agent framework to an avatar in 3D virtual space remains a challenging and open problem, as it means a more refined control over internal psychological processes and external behaviors for multi-dimensional interaction.
%
Moreover, the emergence of human-like behaviors in LLMs has prompted researchers to investigate controllable mental behaviors in LLMs, such as a stable personality~\cite{safdari2023personality} and human simulation in political science~\cite{argyle2023out} and social psychology~\cite{aher2023using}. 
% Thus utilizing the structured prior knowledge from psychological studies to construct a controllable 3D avatar with social intelligence is a new field full of potential but unexplored.
Recently, Social Simulacra~\cite{park2022social} and S$\textsuperscript{3}$~\cite{gao2023s3} build agent systems with autonomous posting and reposting skills in internet community space. Generative Agents~\cite{park23generative} facilitates the formation of social relationships and information diffusion by daily schedules and brief communications within a 2D sandbox gaming space.
\section{Methodology}
\label{sec:method}
\subsection{Text as the Universal Medium}
We define \textit{behavior}, a dictionary-like structured text message to carry key information across modules. For example, \textit{\textbf{$<$speech$>$}Hello! 
\textbf{$<$motion$>$}waves right hand \textbf{$<$place$>$}table} contains pre-set \textit{keys} encapsulated by pointy brackets, followed by the respective \textit{values}, also in natural language. Behaviors are thus interpretable by the LLM and the regular-expression parser. In this work, we focus on \textit{$<$motion$>$}, but we discuss the use of other tokens in the \Supp: \textit{$<$place$>$} triggers navigation and basic state transfer (\eg, sit down), \textit{$<$speech$>$} may be used for face control. 
% We identify three ways to obtain \textit{behavior}: 1) existing work~\cite{story2motion} has shown an LLM can parse a human-written script into a sequence of structured text messages; 2) our Brain (\Sec\ref{sec:method:brain}) generates \textit{behavior} from a psychological process, and 3) our Eye (\Sec\ref{sec:method:eye}) captions the human player in a monocular video to produce \textit{behavior}.

\subsection{Active-Passive Mechanism} 
There exists an intrinsic order in human interaction. For example, ``shaking hands" may appear to be a simultaneous action by two subjects, it typically initiates with one person extending a hand first. Moreover, the other person's action is largely predictable: it is socially appropriate for that person to reciprocate the handshake as a basic courtesy. Drawing from this observation, we design the Active-Passive Mechanism shown in \Fig\ref{fig:method}, where the subject to whom the \textit{behavior} is assigned becomes the ``active" character, whereas the partner becomes the ``passive" character. The active character generates a motion pair for both parties engaged in the interaction, and the passive \textit{behavior} corresponds to the passive motion. Both passive \textit{behavior} and motion are then passed to the passive character. However, the passive character still retains discretion: it only executes the passive motion if its brain ``approves" the passive \textit{behavior} (by prompting the LLM with the suggested behavior and behavior context in its memory). Note that the ``active" and ``passive" roles constantly swap between characters as the interaction progresses.

\subsection{Interactive Motion Synthesis}
\label{sec:method:body}

In our application scenario, the generated actions need to fulfill two main requirements: 1) They must be highly accurate to ensure natural interaction between characters, such as having sufficient contact when shaking hands. 2) They should generate diverse actions to adapt to different plots. In the literature, there are two common solutions for motion synthesis: motion matching and motion generation. However, each of these methods alone can only meet one of our requirements. Therefore, in this paper, we propose a new method called MoMat-MoGen to combine the strengths of both, generating dual-person actions that are both diverse and accurate.
As shown in \Fig\ref{fig:method}, \body leverages motion matching (\Sec\ref{sec:method:body:momat}) to achieve a relevant (but not necessarily perfect) motion from a small database as a prior, and motion generation (\Sec\ref{sec:method:body:mogen}) afterward to diversify the motion with text input while retaining interactive relations between two characters. 

\subsubsection{Motion Matching for High-Quality Motion Prior}
\label{sec:method:body:momat}
% What is motion matching
The motion matching algorithms retrieve motion segments from a database in an auto-regressive manner based on predefined features. 
% Previous motion matching algorithms
The basic motion matching \cite{clavet2016motion} relies on state-based features (e.g., joint position) along with trajectory. The story-to-motion \cite{qing2023storytomotion} further incorporates text-based features to enable semantic control. However, both methods are designed for single-person scenarios. 

% Our work
In this work, we extend the Text-based Motion Matching \cite{qing2023storytomotion} to accommodate interactive scenarios. 
Our objective is to find a pair of motion clips for both characters that align with the query text and trajectory while maintaining a consistent body pose to ensure coherence with the previous motion. In this light, we use a coarse-to-fine motion search strategy, leveraging the text for a high-level semantic understanding of the desired motion, and kinematic features for the low-level control.
%
%-----detailed definition of motion matching:
%Each entry in the database consists of a text label $text_i$ (e.g., "sitting") and a corresponding motion sequence $m_i \in \mathbb{R}^{L \times D}$, where $L$ represents the number of frames and $D$ denotes the dimensionality of each frame, capturing body joint rotation and overall translation.
%It's worth noting that the motion database comprises short clips, and the duration of the target action is provided in the ChatGPT-generated instruction. This flexibility allows the system to scale to arbitrary lengths of motion by retrieving and blending short clips into longer motions.
%At time step $t=0$, an initial motion $m_0$ is randomly selected based on $text_0$ and placed at coordinates $(x_0, y_0)$. At each subsequent time step $t$, given the current position $(x_{t}, y_{t})$, the associated text label $text_{t}$ from the set $\mathcal{S}(t)$, and the previous motion $m_{t-1}$, metrics are utilized to identify the best matching motion from the database.
%
\textbf{First}, we incorporate semantic control by employing a pre-trained sentence encoder \cite{liu2019roberta} to extract text embedding from the query text. Then top-$K_1$ candidates are selected using cosine similarity for subsequent matching. 
%
%-----Trivial trick
%However, two challenges arise: (1) imperfect text matching may result in irrelevant motions, and (2) some datasets may contain a small number of low-quality clips. To address these challenges, we employ outlier removal techniques to discard noisy motion clips that deviate significantly from the distribution center.
%
% basic motion matching
\textbf{Second}, trajectory and coherence constraints are incorporated through joint kinematics features. For the trajectory constraint, the features include the position of the hip joint and the facing direction. For the coherence constraint, the features include positions, velocities, and rotation in 6D space \cite{zhou2019continuity} of the body joints. 
% modification for two-person scenario
For the two-person scenario, a new challenge arises: the interaction between the two characters requires that their relative positions and orientations align with the intended motions. Therefore, the relative position of the other character is taken into account to minimize blending artifacts caused by long-distance movements. 
% how to retrieval
To expedite retrieval, the aforementioned features are pre-calculated and Z-Score normalization is applied to account for magnitude differences. During retrieval, query features are calculated based on the current pose and target trajectory, and the Top-$K_2$ motions are selected using the Euclidean distance. Random selection is used if multiple suitable candidates exist. 

% neural motion blending
Moreover, the neural motion blending model is used \cite{qing2023storytomotion} to generate the transition motion. Hence, the short motion clips are blended into long motions. Notably, the blending model provides smooth transitions to let the character move to the correct place and turn in the correct direction to interact with the other character. Furthermore, motion matching is used for single-person motions. This includes 1) navigation in the scene, where multi-agent path finder~\cite{sharon2015conflict} is used to plan a collision-free trajectory, follow which walking motions are matched from AMASS, and 2) basic character-object interaction such as ``sit down on the chair". More details are included in the \Supp.

\begin{figure*}[t]
    \centering
    \includegraphics[width=\linewidth]{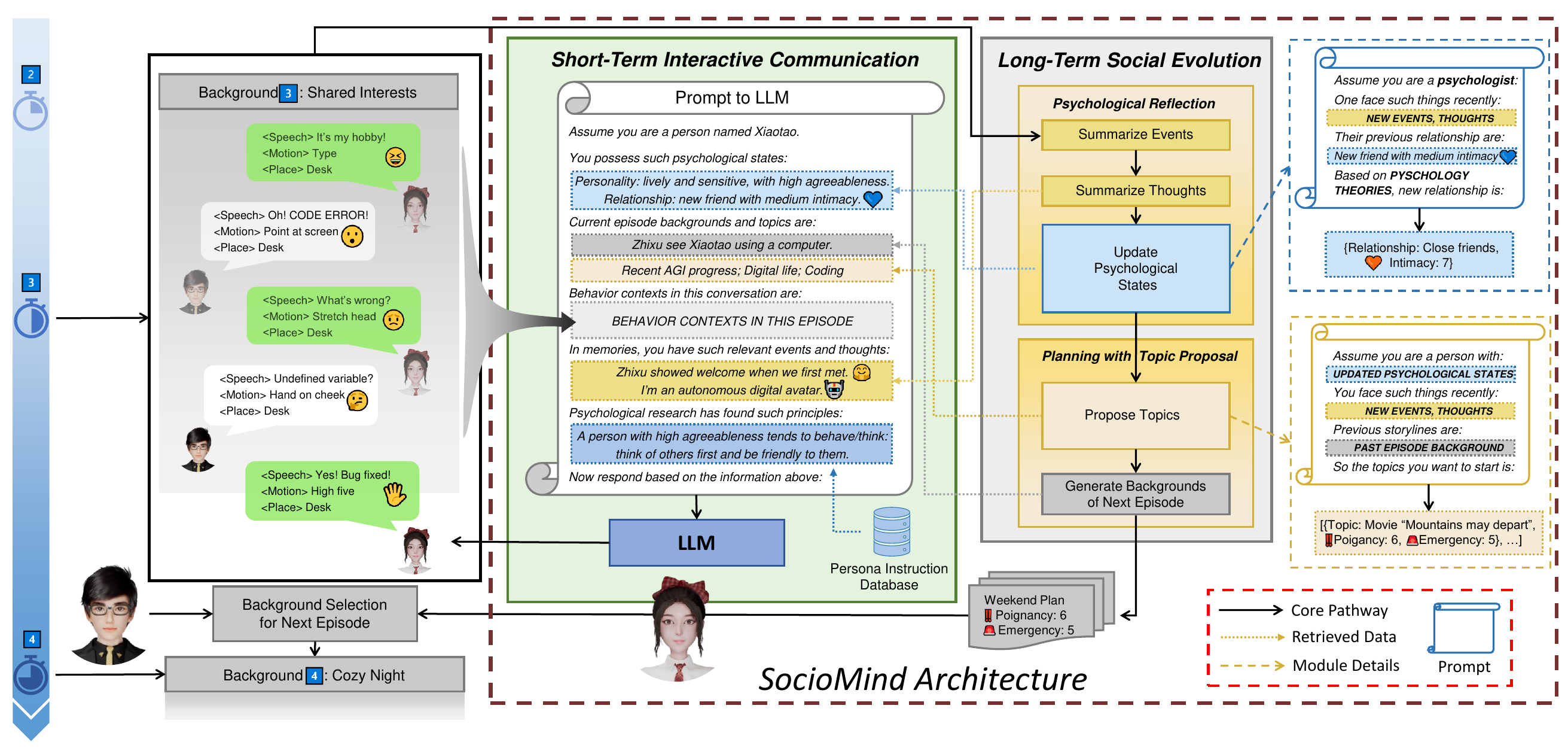}
    \vspace{-20pt}
    \caption{Overview of \textbf{\brain}. To enable 3D characters with social intelligence, our brain utilizes psychological principles to emulate controllable behaviors for short-term interactive communication. For long-term social evolution, our brain assures the consistency of psychological states and plots towards initial settings through psychological reflection and planning with topic proposal. 
    % Zoom in for detailed prompts.
    }
    \label{fig:brain}
    \vspace{-10pt}
\end{figure*}

\subsubsection{Motion Generation for Diversity}
\label{sec:method:body:mogen}
The MoMat-MoGen structure shares many similarities with ReMoDiffuse~\cite{zhang2023remodiffuse}, incorporating retrieval techniques to enhance generation quality. Applying ReMoDiffuse to interaction generation is not trivial. \textbf{Firstly}, it lacks a mechanism for interaction modeling, resulting in a poor correlation between the two generated sequences. \textbf{Secondly}, achieving physical naturalness is challenging if we solely rely on data-driven generation. To address these challenges, we 1) design a Dual-path Semantic-Modulated Attention module (DSMA) to model the interaction between two individuals. 2) During the inference stage, we adaptively extract interaction information from the referenced motion and use it as a constraint for the denoising process, providing additional supervisory signals.

\noindent\textbf{Motion Diffusion Model.} In the diffusion process, it repeatedly adds Gaussian noises to the clean motion sequence pair $(\mathbf{x_0},\mathbf{y_0})$ to noised sequence pair $(\mathbf{x}_T, \mathbf{y}_T)$.
\begin{equation}
    \begin{aligned}
        &q(\mathbf{x}_{T},\mathbf{y}_T \vert \mathbf{x}_0,\mathbf{y_0}) \,:=\, \prod_{t=1}^{T} q(\mathbf{x}_t,\mathbf{y}_t \vert \mathbf{x}_{t-1},\mathbf{y}_{t-1}), \\
        &q(\mathbf{x}_t,\mathbf{y}_t \vert \mathbf{x}_{t-1},\mathbf{y}_{t-1}) \,:=\, \mathcal{N}(\sqrt{1-\beta_t}(\mathbf{x}_{t-1}, \mathbf{y}_{t-1}),  \beta_t\mathbf{I}),
    \end{aligned}
\end{equation}
where $T$ is the total diffusion steps. $\beta_1, \cdots, \beta_T$ is a series of pre-defined variance scales for different timesteps. In the reverse process, given the text prompt $P$, the motion matching result $\bar{\Theta}$ and the timestep $t$, the initial sequence pair is estimated by a network $S_{\theta}(\mathbf{x}_t,\mathbf{y}_t,t,\bar{\Theta}, P)$.

\noindent\textbf{Network Architecture.} Similar to ReMoDiffuse, our network is built upon transformer layers. We modify the design of the attention module in ReMoDiffuse to better capture the interaction. Specifically, in our DSMA module, the input includes motion feature sequences, $f_x$ and $f_y$, feature sequences extracted from the motion matching results, $r_x$ and $r_y$, and text feature sequences $p_x$ and $p_y$. When refining $f_x$, we utilize the generated global attention from $f_x, f_y, r_x, p_x$. The process is similar when refining $f_y$. This approach ensures a more comprehensive fusion of text information, interaction states, and prior information from motion matching.

\noindent\textbf{Training and Inference.} In the training stage, we only use the reconstruction loss as the target:
\begin{equation}
\mathcal{L}=\operatorname{MSE}((\mathbf{x}_0,\mathbf{y}_0), S_{\theta}(\mathbf{x}_t,\mathbf{y}_t,t,\bar{\Theta}, P)).
\end{equation}
In the inference stage, we introduce a contact loss to make the interaction part more natural.
\begin{equation}
\bar{S}=S+\lambda \cdot \nabla(\sum\limits_{i,j_1,j_2} \Arrowvert \bar{D}_{i,j_1,j_2} - D_{i,j_1,j_2} \Arrowvert \cdot [\bar{D}_{i,j_1,j_2} < \gamma]),
\end{equation}
where $\bar{D}_{i,j_1,j_2}$ indicate the distance between the $j_1$-th joint and the $j_2$-th joint in the $i$-th frame from the motion matching results. $D_{i,j_1,j_2}$ is the distance from the motion generation results. $[\cdot]$ is the Iverson bracket whose value is $1$ if and only if the expression inside the parentheses is true. Otherwise the value will be $0$. This auxiliary loss enforces the generated results to imitate the interaction pattern from the prior information and will yield more natural motions.

\subsection{Controllable Emulation of Human Psychology}
\label{sec:method:brain}

From a social psychology perspective, human social intelligence is characterized by 1) various and patterned interactive behaviors during short-term communication~\cite{berger2010handbook}, and 2) the evolution of emotions, attitudes, and relationships \etc over long-term interactions~\cite{cialdini2004social, newcomb1956prediction, rempel1985trust}.
Hence, we propose \brain, a text-centric cognitive framework derived from the idea of ``from strings to symbolic AGI"~\cite{sumers2023cognitive, newell1989symbolic}.
As shown in \cref{fig:brain}, when avatars conduct short-term communication, \brain takes the text interpretation results as input. 
Guided by psychological states, persona instructions, relevant memories, and context behaviors, \brain outputs \textit{behavior} to manipulate the 3D character.
For long-term social evolution, \brain autonomously reflects on psychological states at the end of each interaction session, where several rounds of behaviors are generated between characters. We refer to such a session as a \plot. It also determines the background for the next \plot through planning with topic proposal.

\subsubsection{Short-Term Interactive Communication}
% \paragraph{Behavior definition.}
% Prior research points out that social communication
% % work as the most distinguished ability of human~\cite{levy2004education}, 
%  composed of verbal and non-verbal messages~\cite{berger2010handbook}.
% Non-verbal cues (body language, facial expressions, proxemics, and \etc), play a significant role in scenes such as cross-cultural meeting, education, and romantic encounters~\cite{hall1966hidden, hall1973silent}.

% Our prompt to the LLM consists of four components: psychological states, persona instructions, relevant events and thoughts from the memory system, and behavior contexts in this episode.

% \paragraph{Psychological states.}
Short-term interactive communication means interactive behaviors in multiple dimensions, which are strongly influenced by internal psychological states. 
Here we adopt the most critical dimensions with psychological theories: Big Five Trait model~\cite{john1999big} for personality, 
% PAD model~\cite{schachter1962cognitive} for emotions, 
long-term and short-term motivations~\cite{vallacher1987people}, 
central beliefs~\cite{higgins1987self}, 
and trust~\cite{rempel1985trust, rotter1967new}, intimacy~\cite{newcomb1956prediction}, and supportiveness~\cite{cohen1985stress, cohen2004social} in social relationships. 
However, the safe alignment enables current LLMs with a predefined personality, such as a friendly and cooperative disposition~\cite{openai2023gpt4, touvron2023llama, safdari2023personality}.
To enhance the controllability of psychological states on behaviors, we introduce persona instructions: few-shot exemplars from a reverse-engineering approach on open psychological tests.

% \begin{itemize}
%     \item For personality, we use Big Five Trait model~\cite{john1999big}, which comprises five dimensions: openness (O), conscientiousness (C), extraversion (E), agreeableness (A), and neuroticism (N). Users can provide numerical values for these five dimensions (Likert scale, ranging from 1 to 9) or textual descriptions. The personality is set initially. R
%     \item For emotions, we choose the PAD model~\cite{schachter1962cognitive}, commonly used in body language~\cite{duncan1969nonverbal} and animations~\cite{nishida2008conversational}, consisting of three dimensions: pleasure (how positive or negative), arousal (level of mental alertness and physical activity), dominance (amount of control and influence) in a Likert scale from 1 to 9.
%     \item In terms of motivation, we apply long-term and short-term motivations as propose by Robin \etal~\cite{vallacher1987people}. Long-term motivations are specified in the initial setting.
%     \item For the self, we emphasize central beliefs, reflecting an individual's worldview~\cite{higgins1987self}.
%     \item For social relationships, we introducing three dimensions based on social support theory~\cite{cohen1985stress, cohen2004social}, social trust theory~\cite{rempel1985trust, rotter1967new}, and research on intimate relationships~\cite{newcomb1956prediction}: trust, intimacy, and supportiveness in a Likert scale from 1 to 9. Additionally, we configure an attribute representing the attitude towards others with text description.
% \end{itemize}

\paragraph{Persona instructions.}
In CoT~\cite{wei2022chain}, constructing accurate few-shot exemplars can effectively enhance the reasoning capability of LLMs. 
When prompting LLMs to infer behavior based on human psychological states, crafting precise and reliable exemplars presents a challenging task due to the lack of an exemplar database with high quality. 
Considering that lots of psychological tests~\cite{costa2008revised, hathaway1940multiphasic, myers1985manual, eysenck1975manual} measure psychological traits through observable behavior, we adopt a reverse-engineering method to build a database of trait-to-behavior relationships from psychological tests.
For psychological tests, we choose International Personality Item Pool (IPIP)~\cite{deyoung2007between, du2015using, simms2011computerized}, an open-sourced tool with over 3,000 items and 250 scales for creating advanced measures of personality, motivations, and \etc
Each item, called \textit{persona instruction}, in this database follows the format: 
``A person with \{extend\} \{trait dimension\} tends to behave/think: \{behavior\}",
where \{extend\} are ``high" or ``low" according to the test questionnaire setup.
%During interactive behavior generation, we acquire the few-shot exemplars by retrieving the most similar persona instructions from text embeddings based on psychological traits and ongoing behaviors.
For interactive behavior generation, we retrieve the most similar persona instructions by text embeddings to obtain few-shot exemplars. 

% \paragraph{Relevant memories.}
% Following the framework of the cognitive language agent~\cite{sumers2023cognitive, park23generative} , there are two types of episodic memories within our memory system: `events' and `thoughts'. 
% Events represent occurrences or facts perceived by the agent, whereas thoughts are ideas, musings, or attitudes generated by the agent based on its personality and past experiences. 
% Each event or thought in our memory system has parameters such as poignancy $p_{m}$ (ranging from 1 to 9), text description $D_{m}$, and keywords. 
% While retrieving events or thoughts, we utilize the Ebbinghaus forgetting curve~\cite{ebbinghaus2013memory, averell2011form} to compute the score:
% \begin{align}
%     s(D_{b}, D_{m}) &= cos(\phi(D_{b}), \phi(D_{m})) \times \\
%     &[a + (1-a) \cdot e^{- k \cdot \frac{\Delta T}{2^{N_{m}} \cdot p_{m}}}]
% \end{align}
% , where $cos(\phi(D_{b}), \phi(D_{m}))$ denotes the similarity between the current behavior $b$ and the memory item $m$ (event or thought) using text embedding function $\phi$, $\Delta T$ represents the time elapsed since the last recollection, $N_{m}$ is the accessed times of the item, $a$ and $k$ are hype-parameters.
% We retrieve the top $M$ memory items as prompt input, enabling the avatar to learn interaction strategies from past experiences.

\subsubsection{Long-Term Social Evolution}
Long-term social intelligence requires consistency in two aspects with the initial character setup: 1) the evolution of psychological states such as emotions, relationships, and motivations \etc towards others~\cite{cialdini2004social, newcomb1956prediction, rempel1985trust, cohen2004social}; 2) the progression of overall plots or events~\cite{altman1973social}.
\brain achieves the former aspect through psychological reflection and the latter aspect through planning with the topic proposal.

\paragraph{Psychological Reflection.}
Theories in social cognitive psychology~\cite{heider2013psychology, bandura1977social, chiu1997lay} suggest that humans learn, attribute, and form judgments about others from past experiences. 
Therefore, we introduce a reflection mechanism based on psychological principles. 
Within each \plot, agents introspect on their emotions periodically. 
At the end of each \plot, agents summarize events and their thoughts into a memory system based on the behavior contexts.
Events represent occurrences or facts perceived by the agent, whereas thoughts are ideas, musings, or attitudes generated by the agent based on their personality and past experiences.
Leveraging current events and thoughts, agents retrieve past relevant events and thoughts, and reflect on their motives, central beliefs, and social relationships.
For instance, after \textit{`knowing they share the same interests'}, the \textit{intimacy} of two characters will increase with psychological reflection.

\paragraph{Planning with Topic Proposal.}
% Life-long simulation of 3D avatars is computationally expensive and redundant. Since psychological states and behaviors are influenced by pivotal events~\cite{james2007principles} 
We create a planning module with a topic proposal mechanism for diverse and plausible story progression. 
After each psychological reflection, an agent independently proposes new topics for the next episode based on past memories and character settings. 
Each agent then proposes the background and initial states of both agents for the upcoming episode based on the proposed topics. 
The two agents collect the topics proposed by them and select the most important one for the next episode.
Through this methodology, the two agents can continuously interact with each other from one episode to the next. 
For example, after the topic proposal, the character wants to start several topics (such as the movie \textit{`Mountains may depart'} with the highest \textit{emergency} and \textit{poignancy}) and generate the background \textit{`Weekend Plan'} for next episode. 
The two characters, based on the proposals offered by each, will select an option that holds both high priority and significance, forming the background for the subsequent episode.
% Furthermore, we allow users to add manually set events between episodes to compensate for the limitations of current simulation system, specifically the limited indoor space and two-person setup.

\begin{table*}[ht]
\centering
\caption{\textbf{Interactive Motion Synthesis results on the DLP test set.} `$\uparrow$'(`$\downarrow$') indicates that the values are better if the metric is larger (smaller). We run all the evaluations 20 times and report the average metric and 95\% confidence interval is. The best result are in bold and the second best result are underlined. Our MoMat-MoGen method achieves the best balance between accuracy and diversity.} 
\label{tab:humanml3d}
\vspace{-10pt}
\setlength{\tabcolsep}{2mm}
\resizebox{\textwidth}{!}{
\small
\begin{tabular}{lccccccc}
\hline

\multirow{2}{2cm}{\centering Methods} & \multicolumn{3}{c}{\centering R Precision$\uparrow$} & \multirow{2}{2.0cm}{\centering FID$\downarrow$} & \multirow{2}{2.0cm}{\centering MM Dist$\downarrow$} & \multirow{2}{2.0cm}{\centering Diversity$\uparrow$} & \multirow{2}{2.5cm}{\centering MultiModality$\uparrow$} \\
& Top 1 & Top 2 & Top 3 \\
\hline
Real motions & $0.541^{\pm .002}$ & $0.758^{\pm.002}$ & $0.850^{\pm.002}$ & $0.000^{\pm.000}$ & $3.430^{\pm.012}$ & $4.207^{\pm.071}$ & -\\ 
\hline

MotionDiffuse~\cite{zhang2022motiondiffuse} & $0.035^{\pm.004}$ & $0.058^{\pm.005}$ & $0.098^{\pm.007}$ & $14.883^{\pm.824}$ & $4.199^{\pm.21}$ & $0.677^{\pm.018}$ & $0.655^{\pm.018}$ \\

ReMoDiffuse~\cite{zhang2023remodiffuse} & $0.425^{\pm.002}$ & $0.627^{\pm.003}$ & $0.773^{\pm.003}$ & $0.131^{\pm.004}$ & $3.582^{\pm.015}$ & \underline{$4.097^{\pm.052}$} & \underline{$0.472^{\pm.009}$} \\

% ComMDM & \\

InterGen~\cite{liang2023intergen} & $0.403^{\pm.003}$ & $0.582^{\pm.003}$ & $0.728^{\pm.003}$ & $0.082^{\pm.002}$ & $3.615^{\pm.014}$ & $\mathbf{4.186^{\pm.048}}$ & $\mathbf{0.728^{\pm.021}}$ \\

\hline

Ours (MoMat Only) & $\mathbf{0.517^{\pm.001}}$ & $\mathbf{0.652^{\pm.001}}$ & $\mathbf{0.802^{\pm.001}}$ & $\mathbf{0.034^{\pm.000}}$ & $\mathbf{3.313^{\pm.001}}$ & $0.332^{\pm.001}$ & $0.002^{\pm.000}$ \\

Ours (MoMat-MoGen) & \underline{$0.495^{\pm.003}$} & \underline{$0.651^{\pm.004}$} & \underline{$0.792^{\pm.004}$} & \underline{$0.071^{\pm.002}$} & \underline{$3.561^{\pm.017}$} & $4.025^{\pm.050}$ & $0.452^{\pm.012}$ \\

\hline
\end{tabular}}
\end{table*}
\begin{table*}[ht]
\centering
\caption{\textbf{Motion Captioning results on the HumanML3D test set.} Our evaluation methodology aligns with the TM2T~\cite{guo2022tm2t} metrics, but we uniquely utilize unprocessed ground truth texts for calculating linguistic metrics as done in MotionGPT~\cite{jiang2023motiongpt}.}
\label{tab:motioncaptioningmetrics}
\vspace{-2mm}
\setlength{\tabcolsep}{4mm}
\resizebox{\textwidth}{!}{
\small
\begin{tabular}{lccccccccc}
\hline
\multirow{2}{1.5cm}{\centering Methods} & \multicolumn{2}{c}{\centering R Precision$\uparrow$} & \multirow{2}{1.5cm}{\centering MMDist $\downarrow$} & \multirow{2}{1.5cm}{\centering CIDEr $\uparrow$} & \multirow{2}{1.5cm}{\centering Blue@1 $\uparrow$} & \multirow{2}{1.5cm}{\centering Blue@4 $\uparrow$} & \multirow{2}{1.5cm}{\centering Rouge $\uparrow$} & \multirow{2}{1.7cm}{\centering BertScore $\uparrow$} \\
& Top 1 & Top 3 & & & & & & \\
\hline
Real motions & $0.523$ & $0.828$ & $2.901$ & $-$ & $-$ & $-$ & $-$ & $-$ \\
\hline
TM2T~\cite{guo2022tm2t} & $0.516$ & $0.823$ & $2.935$ & $16.8$ &  $\underline{48.9}$ & $7.00$ &  $\underline{38.1}$ & $32.2$ \\
MotionGPT~\cite{jiang2023motiongpt} &  $\underline{0.543}$ & $\underline{0.827}$ & $\underline{2.821}$ & $\underline{29.2}$ & $48.2$ & $\underline{12.5}$ & $37.4$ & $\underline{32.4}$ \\
\hline
Ours & $\mathbf{0.551}$ & $\mathbf{0.832}$ & $\mathbf{2.813}$ & $\mathbf{36.2}$ & $\mathbf{51.1}$ & $\mathbf{15.5}$ & $\mathbf{41.9}$ & $\mathbf{35.0}$ \\
\hline

% Real & KIT-ML~\cite{plappert2016kit} & $0.399$ & $0.793$ & $2.772$ & $-$ & $-$ & $-$ & $-$ & $-$ \\
% TM2T & KIT-ML & $0.359$ & $0.668$ & \cellcolor{blue!25} $3.298$ & \cellcolor{blue!25} $25.29$ & $36.42$ & \cellcolor{blue!25} $7.98$ & $31.26$ & $20.07$ \\
% MotionGPT & KIT-ML & \cellcolor{blue!25}$0.392$ & \cellcolor{blue!25}$0.723$ & $3.341$ & $12.32$ & \cellcolor{blue!25}$40.51$ & $6.59$ & \cellcolor{blue!25}$38.79$ & \cellcolor{blue!25}$24.50$ \\
% Ours & KIT-ML & \cellcolor{red!25}$0.410$ & \cellcolor{red!25}$0.765$ & \cellcolor{red!25}$2.647$ & \cellcolor{red!25}$71.06$ & \cellcolor{red!25}$53.88$ & \cellcolor{red!25}$22.91$ & \cellcolor{red!25}$50.63$ & \cellcolor{red!25}$46.13$ \\
% \hline
\end{tabular}}
\end{table*}

\subsection{Motion Captioning}
% Inspired by the success of instruction tuning in multi-modal models such as Otter~\cite{li2023otter}, our model replicates similar structural and training strategies, focusing specifically on text-motion interplay. 
We design a motion-to-text translation module inspired by the success of instruction tuning in multi-modal models such as Otter~\cite{li2023otter}, implemented with a retrieval-augmented motion encoder similar to that in ReMoDiffuse~\cite{zhang2023remodiffuse}, with specialized modifications in its attention module to amplify motion feature extraction via multimodal data integration. This encoder is seamlessly integrated with the pretrained MPT-1B RedPajama language model, aligning with structural paradigms in the text-vision domain. 
% This combination not only augments motion feature representation for more effective language guidance but also leverages prior knowledge to advance the motion captioning process. 
Due to space constraints, we include an illustration of the module and more details in the \Supp.
\section{Experiments}
\label{sec:experiments}
To the best of our knowledge, \name is the first comprehensive framework to enable autonomous social characters with articulated 3D bodies. Hence, we break down \name into its modules for a fair comparison with existing arts. 

\subsection{Datasets}
\paragraph{\short-MoCap.}
% To establish a large-scale dataset of high-quality interactive motions, we employ an optical MoCap system consisting of 30 cameras to capture body motions and inertial sensor-based MoCap gloves to track hand motions. We hire four professional actors/actresses (two male and two female), who agree that their motion data can be used for research purposes. The raw Mocap data tracks 53 marker points on each subject's body surface at 120 FPS. We then downsampled and processed them into SMPL-X format. Firstly body shape parameters (beta) were fitted from first frame marker data for actors. Then pose parameters (theta) in each frame were regressed following the pipeline of SOMA~\cite{}. Mapping the original format of hand poses into MANO~\cite{}, body and hand parameters were combined as an SMPL-X data format. \short-MoCap comprises basic motions (22\%), two-person interaction motions (56\%), and short script plays each containing 5-10  of (22\%), giving rise to over nine hours of human motion and over one million frames of annotated motions at 30 FPS.

To establish a high-quality dataset of semantic interactive motions, we used an optical MoCap system consisting of 30 cameras to capture body motions and inertial sensor-based MoCap gloves to track hand motions. We hired four professional actors/actresses (two male and two female) for more expressive motions.
We processed the MoCap data into SMPL-X~\cite{pavlakos2019expressive} format by SOMA~\cite{SOMA:ICCV:2021} on body parameters and MANO~\cite{MANO:SIGGRAPHASIA:2017} mapping for hand parameters.
For accurate semantic labeling on motions, we employed 10 human annotators to label semantic tags and corresponding motion segments based on on-site demonstration videos and MoCap data.
\short-MoCap comprises basic motions (22\%), two-person interaction motions (56\%), and short script plays (22\%), giving rise to over nine hours of human motion at 30 FPS and over 4K text-to-motion pairs.

\begin{figure}[t]
    \centering
    \vspace{-10pt}
    \includegraphics[width=\linewidth]{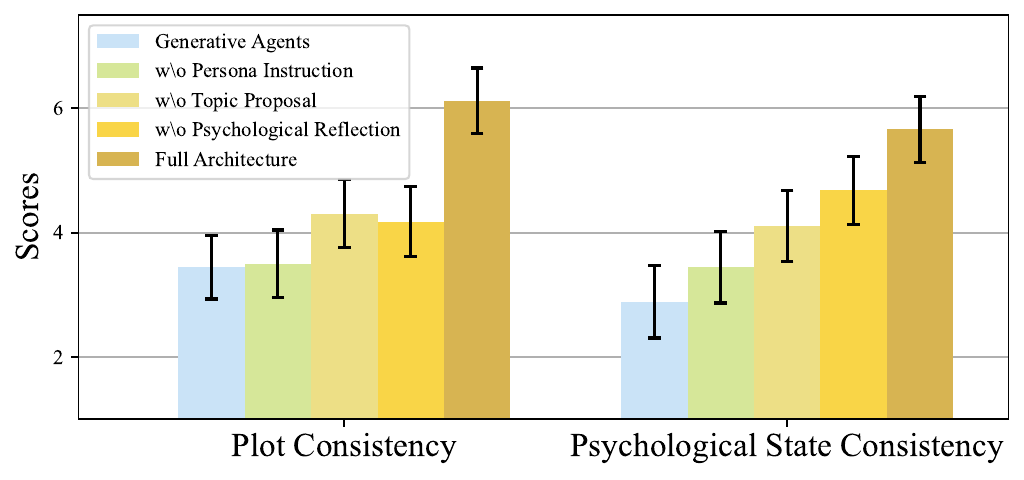}
    \vspace{-20pt}
    \caption{Ablation results on consistency with 95\% confidence.}
    \label{fig:consistency}
    \vspace{-10pt}
\end{figure}

\paragraph{More datasets.} Besides DLP-MoCap, we also evaluated the proposed MoMat-MoGen module on InterHuman~\cite{liang2023intergen}, another public interactive motion dataset. Furthermore, we evaluated the proposed motion captioning module using two key public motion-text datasets: KIT-ML~\cite{plappert2016kit} and HumanML3D~\cite{guo2022generating}. Due to space constraint, we include the test results on KIT-ML and InterHuman in the Supplementary Material.
% We evaluated the proposed motion captioning module using three key public motion-text datasets: KIT-ML~\cite{plappert2016kit}, HumanML3D~\cite{guo2022generating}, and InterHuman~\cite{liang2023intergen}.  Due to space constraint, we include the test results on KIT-ML and InterHuman in the Supplementary Material.

\subsection{Interactive Motion Synthesis}
\label{sec:exp:body}
Table \ref{tab:humanml3d} presents a comparative analysis of our proposed interactive motion generation method against three existing approaches: ReMoDiffuse~\cite{zhang2023remodiffuse}, MotionDiffuse~\cite{zhang2022motiondiffuse}, and InterGen~\cite{liang2023intergen}. Our both methods exhibit significant improvements on the DLP dataset, especially in R precision, FID, and MM Dist metrics. Moreover, our MoMat-MoGen version achieves competitive results in terms of diversity and multi-modality.
% These results demonstrate that our method proficiently generates motions with both high accuracy and outstanding quality.
%Additionally, we have achieved an impressive balance between precision and diversity. This balance is essential for our application, which requires the generated motions to not only closely align with well-matched motion references that possess strong priors, but also be varied and diverse.
Moreover, we have achieved an impressive balance between precision and diversity, which is essential for our application, ensuring that the generated motions closely align with well-matched motion references that possess strong priors while being varied and diverse.

\begin{figure}[t]
    \centering
    \vspace{-10pt}
    \includegraphics[width=\linewidth]{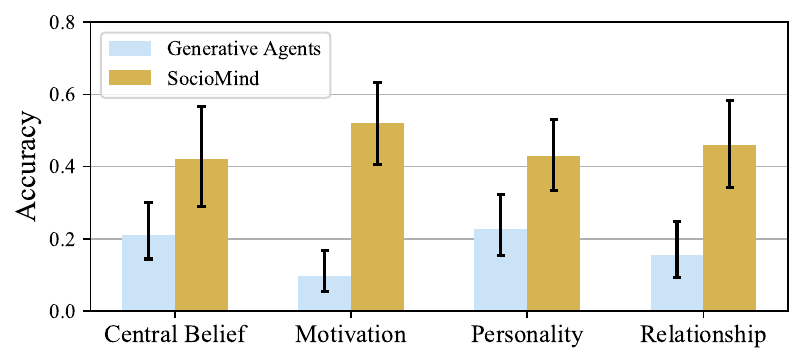}
    \vspace{-20pt}
    \caption{Results on controllibility with 95\% confidence.}
    \label{fig:controllibility}
    \vspace{-10pt}
\end{figure}
\begin{figure*}[t]
    \centering
    \includegraphics[width=\linewidth]{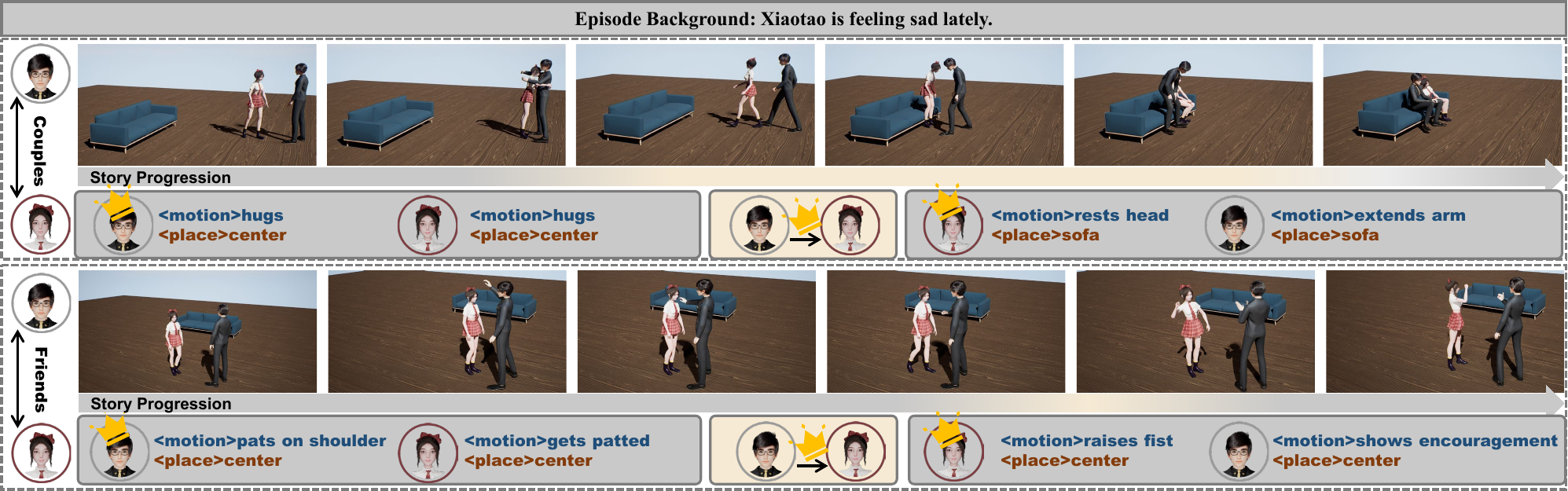}
    \vspace{-18pt}
    \caption{We explore the \textbf{controllability} of \short. Given the same background, manually editing the relationship state between characters, results in different social behaviors. Interestingly, ``couples" tend to have more intimate interactions than  ``friends". The crown indicates the active player. The story progression bar is color-coded in accordance with the stages represented by boxes:  gray boxes represent \textit{behaviors}, whereas yellow boxes represent active-passive swapping in between \textit{behaviors}.}
    \label{fig:visual_society}
\end{figure*}
\begin{figure*}[t]
    \centering
    % \vspace{-10pt}
    \includegraphics[width=\linewidth]{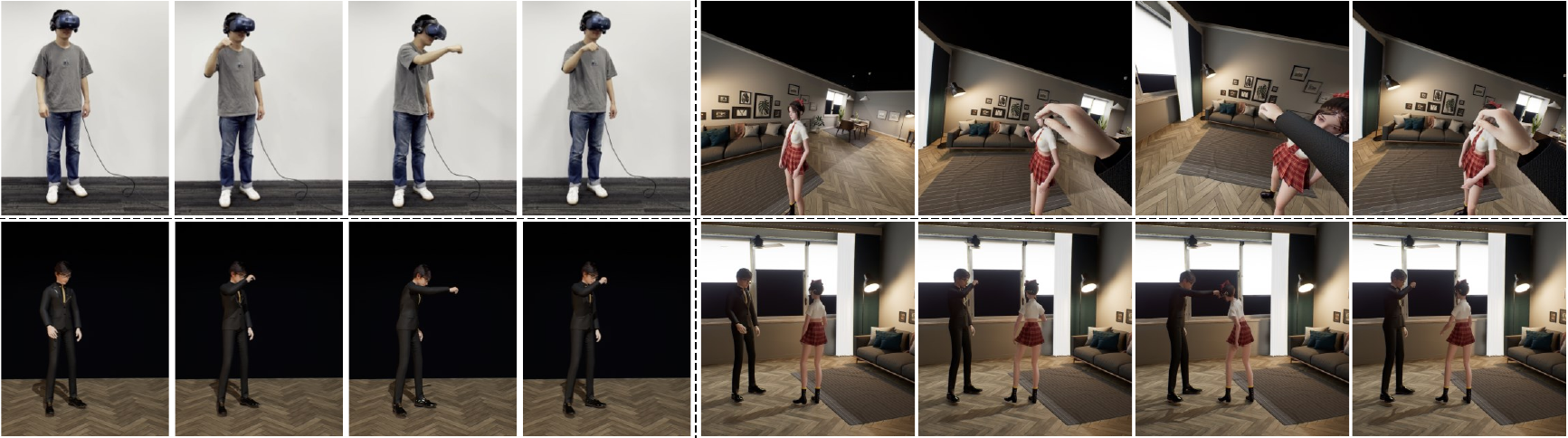}
    \vspace{-18pt}
    \caption{Our motion captioning module translates human motion into text description, allowing a virtual character to respond to the human player's ``fist bump". Top Left: RGB video of the human player; Bottom Left: motion capture~\cite{cai2023smpler} result; Top right: first-person view of the human-driven character; Bottom right: third-person view of the interaction. More details are included in the \Supp.}
    \label{fig:visual_companion}
    % \vspace{-10pt}
\end{figure*}

% [The successful interaction of a fist bump between a real-world player and an autonomous character. ] We captures the player's movements through motion capture, which are then translated into descriptions by the motion captioning module. These descriptions are communicated to an autonomous character, which is controlled by our system combining SocioMind for generating behavioural responses and MoMat-MoGen for generating appropriate physical motions. 

\subsection{Social Intelligence}
\label{sec:exp:brain}

To evaluate the social intelligence of \brain, we measure the controllability of behaviors in short-term interactive communication and the consistency of psychological states and plots in long-term social evolution. 
Following the previous evaluation approach~\cite{park23generative}, we engage 47 human evaluators (ranging in age from 20 to 45 years old) to review the behavioral records of the agents. More details are included in the Supplementary Material.
% We show quantitative results of user studies. 
% For detailed experimental design and qualitative analysis, refer to the Supplementary Material.

\subsubsection{Controllability}
Controllability is measured by whether altering psychological traits can cause noticeable different behaviors in short-term communication. 
We show evaluators the behavioral records of 64 episodes, ask them to select the corresponding psychological traits from the provided options, and subsequently calculate the accuracy.
Results in \cref{fig:controllibility} show that \brain significantly outperforms Generative Agents~\cite{park23generative} in key attributes: central belief, motivation, personality, and relationship, demonstrating the effective guidance of persona instructions for the LLM in simulating interactive human behavior.

\subsubsection{Consistency}
Long-term social evolution consistency implies that the plot development and internal state changes are coherent with initial settings.
To measure this, we use four different types of initial settings (family, crime, romance, and military) to generate records with multiple episodes.
Human evaluators use the records to rate the degrees of consistency on plots and psychological states on a scale of 1 to 9. 
Thus we evaluate the effectiveness of modules in the \brain for social evolution.
Results in \cref{fig:consistency} show that \brain demonstrates superior performance over Generative Agents~\cite{park23generative} on consistency over plots and psychological states, and ablating results show that persona instruction, psychological reflection, and planning with topic proposal are crucial for long-term social evolution.

\subsection{Motion Captioning}
% We conducted a comparative analysis of our proposed methods based on the evaluation metrics mentioned in Section \ref{sec:exp:eye:metrics}, following TM2T~\cite{guo2022tm2t}. Our experimental trials, detailed in Table \ref{tab:motioncaptioningmetrics}, reveal that our model significantly outperforms existing methods, particularly in linguistic metrics, aligning with our design objectives to translate motion sequences into high-quality and precise text. 
% We conducted a comparative analysis of our proposed methods based on the evaluation metrics following TM2T~\cite{guo2022tm2t}. Our experimental trials, detailed in \Tab\ref{tab:motioncaptioningmetrics}, reveal that our model significantly outperforms existing methods, particularly in linguistic metrics, aligning with our design objectives to translate motion sequences into high-quality and precise text. 
Our experiments, following TM2T~\cite{guo2022tm2t}, reveal that our model significantly outperforms existing methods, particularly in linguistic metrics, aligning with our design objectives, the high-quality and precise captioning.

\subsection{Visualization}
\label{sec:exp:visual}
% \input{figures/visual_mogen}
% We demonstrate \name with three applications. First, AI Filmmaker where a manually written or LLM-generated script is turned into a 3D production with virtual actors/actresses. Second, AI Society where characters spontaneously engage in interactions on their wills. Third, AI Companion, where a human is motion-captured and his/her avatar can lead the dialogue with an autonomous character. 
As shown in \Fig\ref{fig:visual_society}, our framework possesses a rational correlation between psychological states and physical behaviors. In addition, our system has the potential to add human players in the virtual world to interact with the digital avatars (\Fig\ref{fig:visual_companion}).
\section{Conclusion}
\label{sec:conclusion}
In this paper, we introduce \name, an innovative and holistic system that harnesses the latest advancements in generative models to create autonomous 3D characters. \short integrates \brain, a text-centric cognitive framework that simulates sophisticated internal psychological processes, and \body, a text-driven motion synthesis pipeline that replicates diverse external physical behaviors. Both modules achieve state-of-the-art performance in the respective domains, enabling the entire system to engage in natural interactions with social intelligence.

\section*{Acknowledgement}
We extend our sincere gratitude to Fei Wang, Jingyi Jiang, Jinyun Lyu, Wanying Zu, Bo Li, Yuanhan Zhang,  Jiawei Ren, and Cunjun Yu for their insights and assistance in the areas of multi-modal models, psychological processes, aesthetics, and the overall framework.

{
    \small
    \bibliographystyle{ieeenat_fullname}
    \bibliography{main}
}

\clearpage
\maketitlesupplementary
\appendix

\section{Overview of the \Supp}

We provide more details of \short, such as the method (\cref{sec:supp:dlp}), the \short-MoCap dataset (\cref{sec:supp:dlp-mocap}), the experiments (\cref{sec:supp:exp}) and additional discussions (\cref{sec:supp:discussion}). 

\paragraph{Demo Video.} A video demonstration is attached as a part of the \Supp. In addition to ``$<$motion$>$", we explore using the ``$<$speech$>$" token to generate audio with OpenAI's text-to-speech engine, followed by Talkshow~\cite{yi2023generating} to synthesize face movements.
\section{Additional Details of \short}
\label{sec:supp:dlp}

\subsection{Scene}
We aim to build 3D characters that articulate with 3D body motions in a 3D scene. In this work, we design our scene to achieve a diverse combination of relative interaction poses: 1) the \textit{center} or \textit{bookshelf} for the cases when both characters are standing (standing-standing interaction), 2) a \textit{sofa} for side-by-side seated interaction, 3) a \textit{dining table} for face-to-face seated interaction, and 4) a computer \textit{desk} for standing-seated interaction. For each piece of interactable furniture, we designed \textit{spots} with positions and facings to guide navigation and human-scene interaction. Note that we focus on human-human interaction in this work and substantially simplify the human-scene interaction to basic ones such as ``sitting down on a chair" or ``standing up from a sofa".

\subsection{Movement Synchronization}
\begin{figure*}[t]
    \centering
    \includegraphics[width=\linewidth]{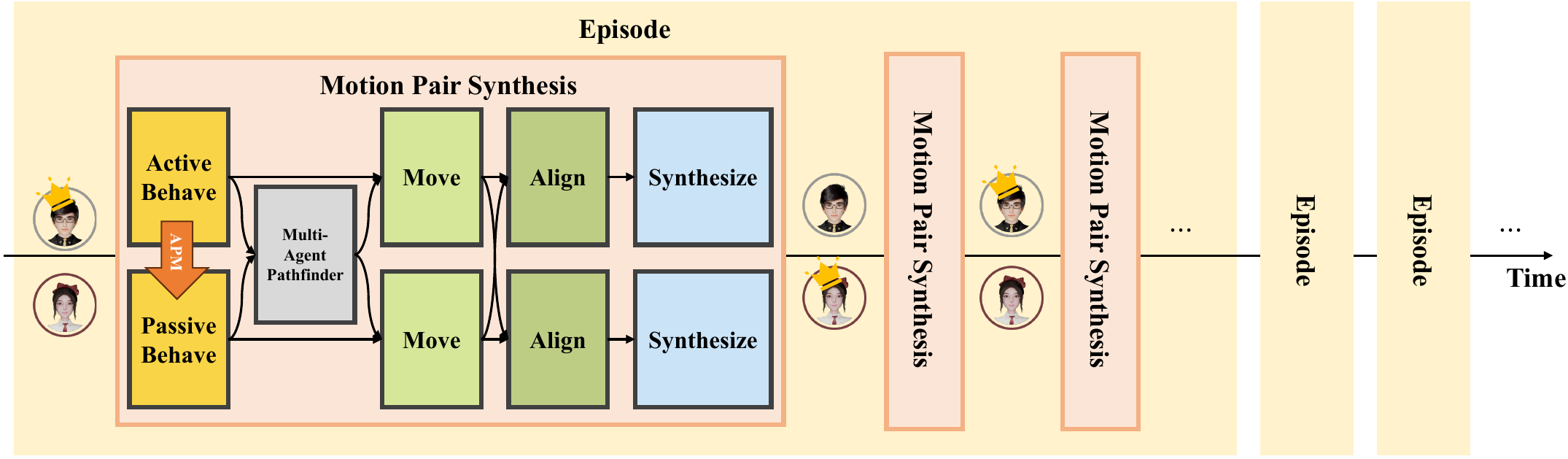}
    \caption{\textbf{Movement synchronization.} Each \textit{behavior} results in an interactive motion pair, that consists of individual movements in the scene and interaction between two characters. To allow characters to navigate in the scene while maintaining synchronization, there are four stages in the motion synthesis process. The crown indicates the active character status, which may or may not swap between characters. APM stands for Active-Passive Mechanism. \cref{fig:method} depicts mainly the \textit{behave} stage.}
    \label{fig:motion_sync}
\end{figure*}
A significant challenge of extending single human motion synthesis to interactive motion synthesis is coordinating multiple characters. Although our Active-passive Mechanism ensures the interaction is naturally aligned through retrieving from a curated motion database, we need to accommodate potential mismatches in motion lengths when the characters navigate from one place in the scene to another. As shown in \cref{fig:motion_sync}, each interaction (interactive motion pair) requires four steps to synchronize the character's motions. 
\begin{itemize}
    \item \textbf{Behave}. In this stage, the active character generates active \textit{behavior} from its \brain and synthesizes both active and passive motions with \body. The passive motion is passed to the passive character. Since the \textit{behavior} contains ``$<$place$>$" token for the location information (\eg, desk or sofa), both characters will register their current position as the starting position with the multi-agent pathfinder~\cite{sharon2015conflict}. Note that at this stage, there are no paths planned or actual motion executed yet.
    \item \textbf{Move}. The multi-agent pathfinder computes collision-free trajectories for both characters. Motion matching is used to synthesize the walking paths if there is a location change. For a more complicated case that involves a change in the basic motion state (\eg, from a standing pose in \textit{center} to a seated pose in \textit{sofa}), there will be additional motion inserted in front (\eg, standing up) or after (\eg, sitting down) the walking trajectory. 
    \item \textbf{Align}. Depending on the starting positions of the characters, the movements typically take a different number of frames for each character. Hence, an alignment is conducted: two characters communicate with each other about their trajectory, and the character with a shorter trajectory has a filler motion (an idle motion) inserted at the end of its trajectory, depending on its basic state (seated or standing).
    \item \textbf{Synthesize}. Movement motions are concatenated with the interaction motions. Motion blending is applied to ensure a smooth transition between motion clips. 
\end{itemize}

\subsection{Retargeting}
In this work, the proposed MoMat-MoGen module could produce high-quality motions in SMPL-X \cite{pavlakos2019expressive} format. To better demonstrate the physical and mental interaction between our social agents in an immersive simulation scenario, two rigged characters are used: a male named Zhixu and a female named Xiaotao. The synthesized motions are retargeted to the target avatar in Blender using a widely used retargeting tool Auto-Rig Pro \cite{autorigpro}. In order to bridge the gap between different skeletons, the bone mapping between different structures is manually configured for the best performance. In addition, we rescale the target avatars to have the identical height as SMPL-X models to avoid any noticeable foot skating and preserve body contact for the interaction between the two characters. Note that the retargeting pipeline can be extended to more characters in the future.

\subsection{Motion Matching}

Our motion matching process consists of two steps. Firstly, we incorporate semantic information by utilizing a pre-trained LLM \cite{liu2019roberta} to extract a text embedding $f_t \in \mathbb{R}^{1024}$ for the query text. Secondly, to enhance coherence and alignment with the query trajectory, we incorporate kinematics features. Specifically, for a motion with $k$ frames, the kinematics features are defined as $x = \{ \textbf{t} \ \textbf{f} \ \textbf{b} \ \textbf{h} \ \textbf{p} \} \in \mathbb{R}^{5k+193}$, where $\textbf{t}\in \mathbb{R}^{2k}$ represents the trajectory position projected on the ground, $\textbf{f}\in \mathbb{R}^{3k}$ denotes the facing direction, $\textbf{b}\in \mathbb{R}^{189}$ represents the 6D space rotation and the position of 21 joints, $\textbf{h}\in \mathbb{R}$ indicates the hip height, and $\textbf{p}\in \mathbb{R}^3$ represents the relative position of other characters. The trajectory positions \textbf{t} and facing \textbf{f} align the resulting motion with the query trajectory, the body-pose features \textbf{b} improve body-pose coherence with a higher emphasis on foot weighting, the hip height \textbf{h} distinguishes seated motion from standing motion, and the relative position \textbf{p} aligns with other characters. During motion matching, the query trajectory obtained from the path-finding algorithm is used to calculate trajectory similarity $\mathcal{T}$ and facing similarity $\mathcal{F}$, while the current pose is used to calculate body-pose similarity $\mathcal{B}$, hip similarity $\mathcal{H}$, and relative position similarity $\mathcal{P}$. Euclidean distance is used for all similarities except for facing similarity, which employs cosine distance. These similarities are normalized using Z-score normalization, and the final similarity $\mathcal{S}$ is obtained as a weighted sum of these similarities.

\subsection{Motion Generation}

\subsubsection{Architecture Details}
\begin{figure}[h]
    \centering
    \includegraphics[width=\linewidth]{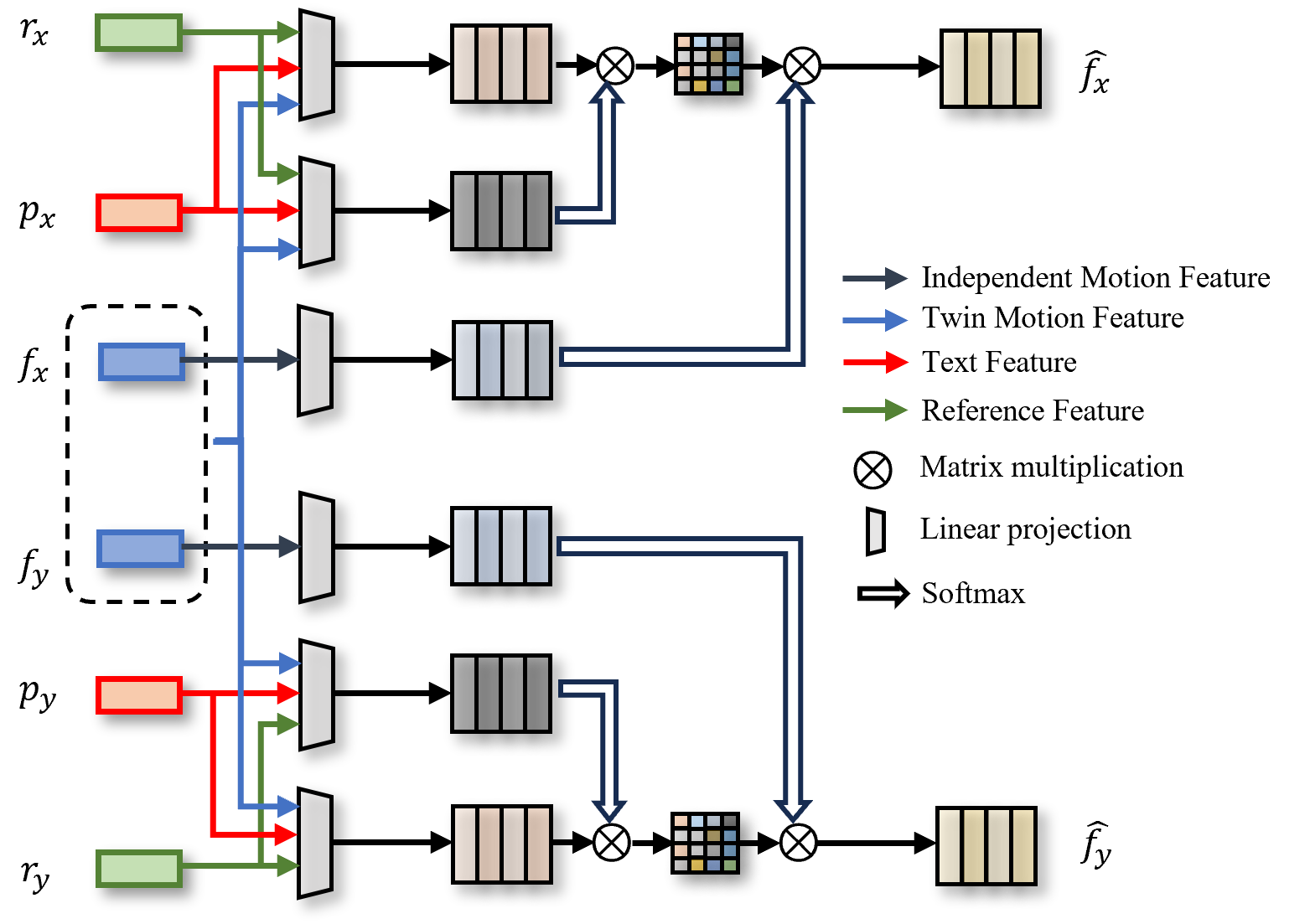}
    \caption{Architecture Detail of \textbf{Dual Semantic-Modulated Attention (DSMA)} module.}
    \label{fig:dsma}
\end{figure}

Our proposed Dual Semantic-Modulated Attention (DSMA) module is built upon the SMA module in ReMoDiffuse~\cite{zhang2023remodiffuse}. The major difference is the introduction of motion interaction. The architecture is shown in \Fig\ref{fig:dsma}. To get the refined feature $\widehat{f}_x$ , we integrate four sources of features: 1) independent motion feature motion $f_x$ from the same sequence; 2) twin motion feature $f_y$ from the partner sequence; 3) text feature $p_x$ from the given description $P_x$; 4) reference feature $r_x$ from the motion matching results $M_x$. In addition, We share parameter weights of DSMA modules and FFN modules to process both actors' motion sequences.

\definecolor{psycho}{HTML}{4169E1}
\definecolor{background}{HTML}{228B22}
\definecolor{topic}{HTML}{FF1493}
\definecolor{behavior}{HTML}{A52A2A}
\definecolor{events}{HTML}{FF4500}
\definecolor{persona}{HTML}{9C27B0}

\subsection{\brain}
In this section, we illustrate the details of each module in \brain{}, along with the corresponding prompt templates. 
We introduce the definition of psychological states in \cref{supp: B psycho states}, the construction of persona instructions in \cref{supp: B persona instructions}, events and thoughts within the memory system in \cref{supp: B events thoughts}, prompts for short-term communication in \cref{supp: B short-term}, psychological reflections in \cref{supp: B psycho reflection}, and the details of the topic proposal mechanism in \cref{supp: B topic proposal}.

\subsubsection{Psychological States}
\label{supp: B psycho states}
Determining whether a 3D character qualifies as a digital life with social intelligence remains an open problem. 
From a psychological perspective, humans are composed of internal psychological processes (mind, such as thoughts, emotions, \etc) and external behaviors~\cite{james2007principles}. Through prolonged studies on the association between internal processes and external behaviors, psychologists have developed various theories, including Big Five Trait~\cite{john1999big} on personality, PAD model~\cite{schachter1962cognitive, mehrabian1980basic} on emotion, hierarchy of needs~\cite{maslow1958dynamic} and long-short term theory~\cite{vallacher1987people} on motivation, self-schema~\cite{higgins1987self} on core self, episodic and semantic system~\cite{buckner2008brain} on memory, attitude~\cite{cialdini2004social}, intimacy~\cite{newcomb1956prediction}, supportiveness~\cite{cohen1985stress, cohen2004social} and trust~\cite{rempel1985trust, rotter1967new} on social relationships.
Herein, we introduce the psychological states in our \brain{}, enabling the simulation of controllable human communicative behaviors.
The psychological states are as follows:

\begin{itemize}
    \item For personality, we use Big Five Trait model~\cite{john1999big}, which comprises five dimensions: openness (O), conscientiousness (C), extraversion (E), agreeableness (A), and neuroticism (N). Users can provide numerical values for these five dimensions (Likert scale, ranging from 1 to 9) or textual descriptions. The personality is set initially.
    \item For emotions, we choose the PAD model~\cite{schachter1962cognitive}, commonly used in body language~\cite{duncan1969nonverbal} and animations~\cite{nishida2008conversational}, consisting of three dimensions: pleasure (how positive or negative), arousal (level of mental alertness and physical activity), dominance (amount of control and influence) in a Likert scale from 1 to 9.
    \item In terms of motivation, we apply long-term and short-term motivations as propose by Robin \etal~\cite{vallacher1987people}. Long-term motivations are specified in the initial setting.
    \item For the self, we emphasize central beliefs, reflecting an individual's worldview~\cite{higgins1987self}.
    \item For social relationships, we introducing three dimensions based on social support theory~\cite{cohen1985stress, cohen2004social}, social trust theory~\cite{rempel1985trust, rotter1967new}, and research on intimate relationships~\cite{newcomb1956prediction}: trust, intimacy, and supportiveness in a Likert scale from 1 to 9. Additionally, we configure an attribute representing the attitude towards others with text description.
\end{itemize}

\begin{figure}[h]
    \centering
    \includegraphics[width=\linewidth]{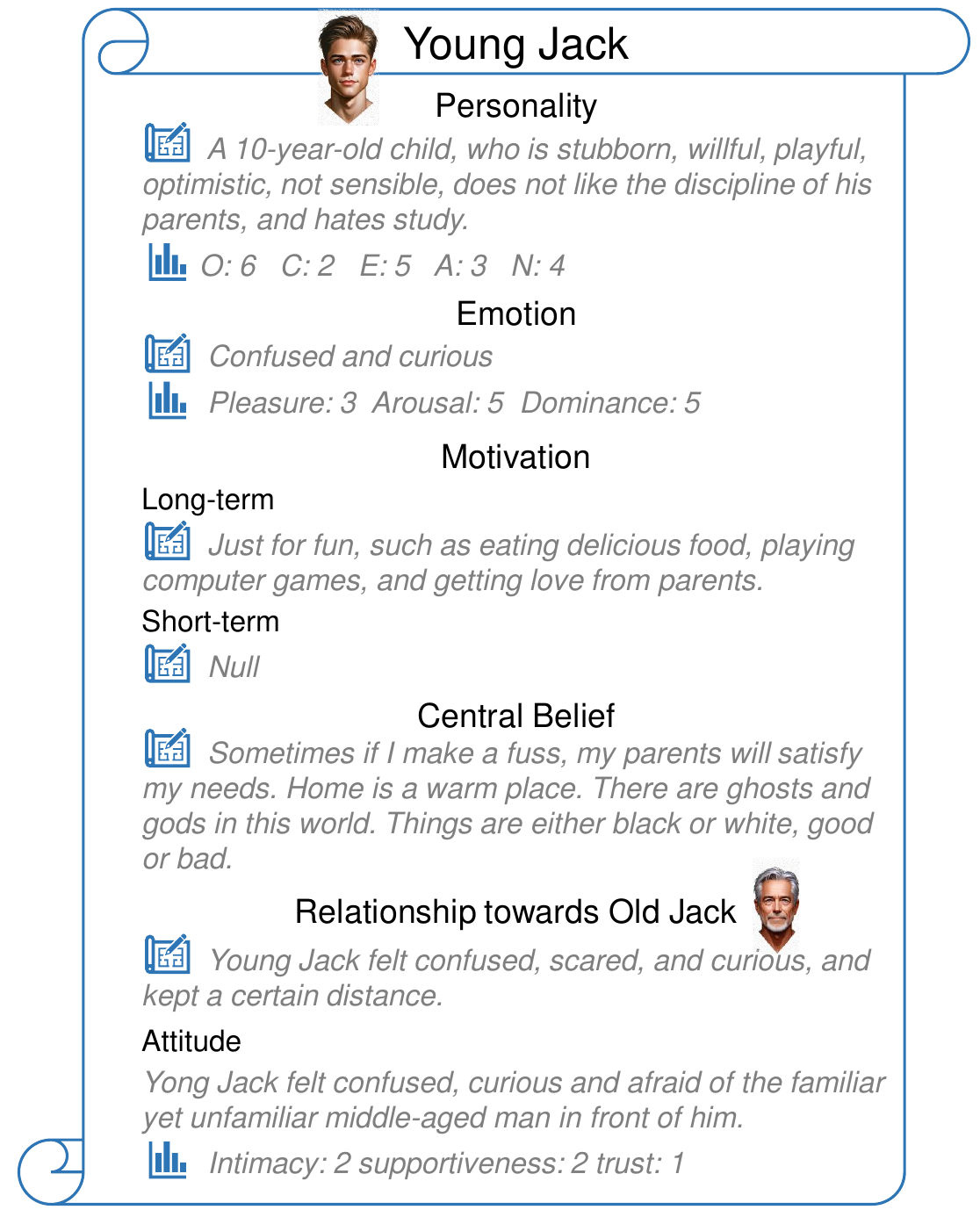}
    \caption{An example of the psychological states of an autonomous character, Young Jack. The background is `\textit{Old Jack traverses through time and engages in communication with his younger self.}'
    }
    \label{fig:B_psycho_states}
\end{figure}

\cref{fig:B_psycho_states} is an example of a character's psychological states.
All psychological states can be defined in textual form, wherein personality, emotions, and social relationships can also be delineated through quantifiable values.
Users can define a character's personality, emotions, and social relationships either through text or by adjusting the numerical values of corresponding variable dimensions. 
In subsequent reasoning based on LLM, both text and numerical descriptions serve as inputs to the LLM as prompts. 
The transformation between text and numerical values is facilitated through LLM. 
For instance, we use the LLM to translate the numerical values of emotion dimensions into text descriptions based on the PAD~\cite{schachter1962cognitive} model. The form of the prompt is as follows:
\begin{tcolorbox}[title={Prompt for translating numerical values into text}]
\textit{Assume you are a very professional psychologist.}\\
\textit{This is a quantitative evaluation of a person:} \textbf{[pleasure: 3, arousal: 5, dominance: 4]}.
\textit{The evaluation based on the PAD theory of psychology and Likert scale (1-9), score on pleasure, arousal and dominance.
Based on the score, describe the emotion of the person.
According to Paul Ekman's basic emotion theory, human have basic emotions: wrath, grossness, fear, joy, loneliness, shock, amusement, contempt, contentment, embarrassment, excitement, guilt, pride in achievement, relief, satisfaction, sensory pleasure, and shame.} \\
\textit{So the description should be:}
\end{tcolorbox}

And the result may be:
\begin{tcolorbox}[title={Output}]
\textit{Mild dissatisfaction or discontentment, coupled with a sense of alertness but not empowerment.}
\end{tcolorbox}

\begin{figure}[t]
    \centering
    \includegraphics[width=\linewidth]{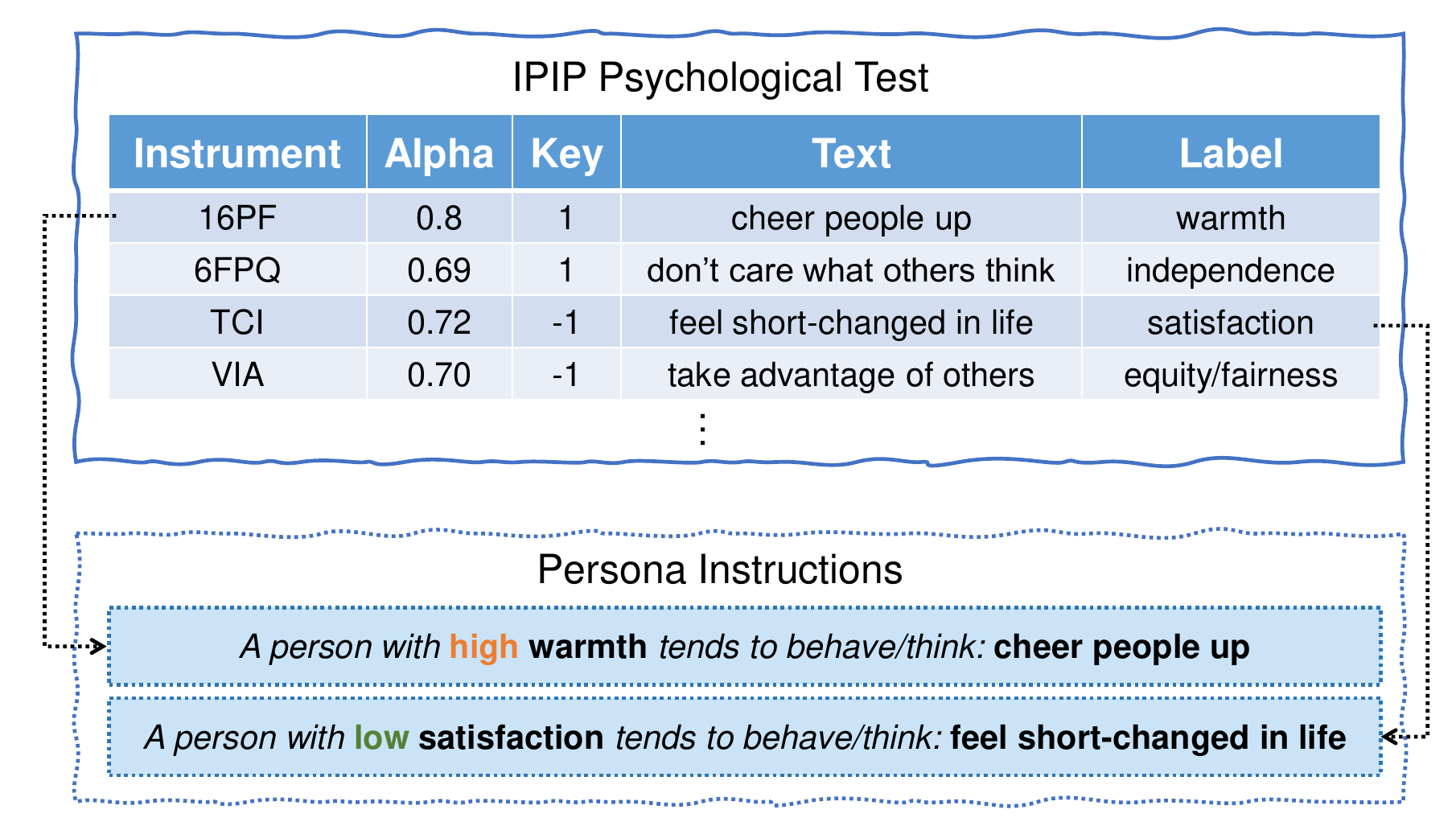}
    \caption{Illustration of our reverse engineering to transfer the psychological test into persona instructions.
    In the table of IPIP, `instrument' is the name of the personality inventory, `alpha' is the Cronbach alpha reliability~\cite{tavakol2011making}, `key' is the keyed direction (+1 for positive and -1 for negative) of the item and its associated construct, the text is the behavior item, and `label' is the trait dimension.
    }
    \label{fig:B_persona_instruction}
\end{figure}

\subsubsection{Persona Instructions}
\label{supp: B persona instructions}
To enhance the LLM's capability in reasoning the coherent association between internal psychological processes and external behaviors, we design a reverse engineering to construct the \textit{persona instruction database} from psychological tests.
Persona instructions serves as few-shot exemplars for Chain of Thought (CoT) reasoning to generate plausible behaviors aligned with human expectations.
In International Personality Item Pool (IPIP)~\cite{deyoung2007between, du2015using, simms2011computerized}, an open-source psychological tool, its original data illustrate how external behaviors can be translated into quantifiable psychological metrics. 
As shown in \cref{fig:B_persona_instruction}, through reverse engineering, we represent each item in IPIP as a persona instruction: 
``A person with {extend} {trait dimension} tends to behave/think: {behavior}",
where extend is `high' when key is 1, and `low' when key is -1.
In short-term communication, we retrieve the most relevant persona instructions by the similarity of the text embeddings from function $\phi$~\cite{textembedding} with current behaviors and psychological states.

\subsubsection{Events \& Thoughts}
\label{supp: B events thoughts}
Following the framework of the cognitive language agent~\cite{sumers2023cognitive, park23generative} , there are two types of episodic memories within our memory system: `events' and `thoughts'. 
Events represent occurrences or facts perceived by the agent, whereas thoughts are ideas, musings, or attitudes generated by the agent based on its personality and past experiences. 
The memory system of Generative Agents becomes larger with time, which poses challenges for retrieving the most relevant events and thoughts~\cite{park23generative}.
Inspired by human memory system~\cite{ebbinghaus2013memory}, we introduce mechanisms for memory reinforcement and forgetting to alleviate this issue.
Each event or thought in our memory system has parameters such as poignancy $p_{m}$ (ranging from 1 to 9), text description $D_{m}$, keywords, and the accessed times $N_{m}$.
To ensure the efficiency of our memory system, we introduce a forgetting mechanism. Initially, we utilize the Ebbinghaus forgetting curve~\cite{ebbinghaus2013memory, averell2011form} to calculate the forgetting rate $r$:
\begin{align}
    r = [a + (1-a) \cdot e^{- k \cdot \frac{\Delta T}{2^{N_{m}} \cdot p_{m}}}],
\end{align}
where $\Delta T$ represents the time elapsed (the number of intervening episodes) since the last recollection, $a$ and $k$ are hype-parameters.
% In our experiments, $a$ is 0.4 for events and 0.1 for thoughts, $k$ is 4 for events and 2 for thoughts. (0.6 for en event and 0.3 for a thought)
When $r$ falls below the threshold $T_f$ , the event or thought will be forgotten and not retrieved.

During memory retrieval, we combine content similarity and the forgetting rate to compute the final score:
\begin{align}
    s(D_{b}, D_{m}) &= cos(\phi(D_{b}), \phi(D_{m})) \cdot r
\end{align}
where $cos(\phi(D_{b}), \phi(D_{m}))$ denotes the similarity between the current behavior $b$ and the memory item $m$ (event or thought) using text embedding function $\phi$~\cite{textembedding}.
We retrieve the top $M$ memory items as prompt input, enabling the avatar to learn interaction strategies from past experiences.

\begin{figure*}[t]

\begin{tcolorbox}[title={Prompt for generating behaviors}]
\textit{\{``role": ``system", ``content": ``Let's do a role play. Assume you are a person named} Young Jack (SELF NAME). \textit{I'm a person with name} Old Jack (SELF NAME).\\
\textit{In this episode, you have such psychological states: }\\
\textcolor{psycho}{\textit{Personality: stubborn, willful, playful, ...;
Emotion: confused and curious, pleasure: 3, ...;
Motivation: just for fun, such as eating delicious food, ...;
...}(PSYCHOLOGICAL STATES)}
\\
\textit{The episode background is: }\\
\textcolor{background}{\textit{Old Jack decides to share his regrets with Young Jack about not being able to spend more time with his mother before she passed away.}
(BACKGROUND)}\\
\textit{The topics you want to start are: }\\
\textcolor{topic}{\textit{regret of limited time with mother.}
(TOPICS)}\\
\textit{Now start our conversation."\}}
\\
\textcolor{behavior}{
\textit{\{``role": ``assistant", ``content": ``\textless{}speech\textgreater{}
Hello!\textless{}motion\textgreater{}waving hands\textless{}place\textgreater{}sofa"\}}\\
\textit{\{``role": ``user", ``content": ``\textless{}speech\textgreater{}
Hi!\textless{}motion\textgreater{}stand up\textless{}place\textgreater{}sofa"\}}\\
 ......\\
 \textit{\{``role": ``user", ``content": ``
\textless{}Memories hold joy and pain.\textgreater{}
Hi!\textless{}motion\textgreater{}lean back against the bookshelf\textless{}place\textgreater{}bookshelf"\}}\\
(BEHAVIOR CONTEXT)}\\
\\
\textit{\{``role": ``user", ``content": ``You have such relevant memories: }\\
\textcolor{events}{\textit{Mom told Young Jack there were ghosts;The old man wanted to share photos with him.}(RELEVANT EVENTS \& THOUGHTS)}\\
\textit{Psychological research has found such principles: }\\
\textcolor{persona}{\textit{A person with high curiosity tends to behave/think: seek explanations of things.}(PERSONA INSTRUCTIONS)}\\
\textit{Now you have two options: end the conversation by output `END' or respond based on the information above. So your reaction is:}

\end{tcolorbox}

    \caption{Prompts for generating interactive behaviors in short-term communication. The use of ALL CAPITAL LETTERS signifies the reference to the complete textual description corresponding to the respective names in this paper.}
    \label{sig:supp: prompt short-term}
\end{figure*}

\subsubsection{Short-Term Communication Generation}
\label{supp: B short-term}
Human responses to external stimuli are influenced on one hand by external environmental factors (episode background, interactive behavior contexts), and on the other hand by internal factors (psychological states, relevant memories, and topics). The correlation between behavior and psychology also follows certain patterns (persona instructions). 
Therefore, to generate interactive behaviors autonomously, we feed LLMs with these factors for reasoning human-like behaviors. 
The specific prompt is shown in \cref{sig:supp: prompt short-term}.
In this paper, the use of ALL CAPITAL LETTERS signifies the reference to the complete textual description corresponding to the respective names.
And a possible response from LLM is:

\begin{tcolorbox}[title={Output}]
\textit{\textless{}speech\textgreater{}
Old Jack, do you think that memories are like ghosts? They can't be seen but they linger with us, bringing both joy and sadness.\textless{}motion\textgreater{}Tracing fingers on bookshelf\textless{}place\textgreater{}bookshelf}\\
\end{tcolorbox}

The format of the output results may not align with the target format. 
In such instances, we use LLMs to reformat the output.

\subsubsection{Psychological Reflection}
\label{supp: B psycho reflection}

\begin{tcolorbox}[title={Prompt for summarizing events}]
\textit{Assume you are a person named} NAME.
\textit{You have such psychological states:} \\
\textcolor{psycho}{PSYCHOLOGICAL STATES}\\
\textit{Psychological research has found such principles: } \\
\textcolor{persona}{PERSONA INSTRUCTIONS}\\
\textit{Now you have a conversation and the contexts are as follow:} \\
\textcolor{behavior}{BEHAVIOR CONTEXTS}\\
\textit{Based on the dialog above, summarize the key events and output the event list.}
\end{tcolorbox}

\noindent
Psychological research reveals that humans possess a reflection system, wherein the brain learns from past events, gradually altering its beliefs and attitudes towards others~\cite{cialdini2004social,  rempel1985trust, cohen2004social, altman1973social}.
Although Generative Agents~\cite{park23generative} adopts a reflection mechanism, it does not adequately consider the multiple factors on a character's long-term social intelligence, such as attitudes, intimacy, beliefs, and motivations. 
Consequently, it can not finely simulate the social evolution process between two characters.
Therefore, we have developed a hierarchical system of reflection to mimic psychological processes. 
After each episode, we use LLMs to summarize the events from the communication (Prompt for summarizing events). One example of the output is:

\begin{tcolorbox}[title={Output}]
\textit{[\{ ``description": ``Young Jack asks Old Jack about the bookshelf and shows curiosity about the memories", ``keywords": [``bookshelf", ``curiosity", ``memories"], ``poignancy": 7, ``emergency": 4\}, ...]}
\end{tcolorbox}

Then the brain generates its own new thoughts based on current events and relevant past events and thoughts.
The structure of `thought' is fundamentally similar to that of `event' in our memory system, with the key distinction lying in its integration of past events and thoughts to generate new thoughts.
Based on these events and thoughts, the brain update its motivations, central belief, and social relationships, thus resulting in social evolution on internal states.
For instance, the brain updates the social relationship with prompt:

\begin{tcolorbox}[title={Prompt for summarizing events}]
\textit{Assume you are a very professional psychologist. Here is a person named} NAME.
\textit{He/She has such psychological states:} \\
\textcolor{psycho}{PSYCHOLOGICAL STATES}\\
\textit{Psychological research has found such principles: } \\
\textcolor{persona}{PERSONA INSTRUCTIONS}\\
\textit{Recent he/she have come across these events and the following thoughts have arisen:} \\
\textcolor{events}{EVENTS \& THOUGHTS}\\
\textit{His/Her previous relationship with } PARTNER NAME \textit{is}: \textcolor{psycho}{SOCIAL RELATIONSHIP}\\
\textit{Output the relationship according to social psychological theory in three dimensions: trust, intimacy, and supportiveness.}
\end{tcolorbox}

The updated social relationship would be:

\begin{tcolorbox}[title={Output}]
\textit{\{ ``description": ``They are getting to know each other", ``intimacy": 3, ``trust": 4, ``supportiveness": 2, ``attitude": ``curious"\}}
\end{tcolorbox}

\subsubsection{Planning with Topic Proposal}
\label{supp: B topic proposal}
To build the long-term evolution of external behaviors, we propose a planning module with topic proposal mechanism to promote the development of storylines. 
This approach is partly inspired by the psychological evolution often linked to significant events~\cite{james2007principles} and partly by techniques used in movies and dramas to progress narratives. Initially, the brain correlates past experiences with newly occurring events to propose topics for the next scene.
At this stage, we also allow users to manually incorporate events not generated within the interaction, such as fragments of past memories and contemporary news events.

\begin{tcolorbox}[title={Prompt for proposing topics}]
\textit{Here is a person named} NAME.
\textit{He/She has such psychological states:} \\
\textcolor{psycho}{PSYCHOLOGICAL STATES}\\
\textit{Psychological research has found such principles: } \\
\textcolor{psycho}{PERSONA INSTRUCTIONS}\\
\textit{The person has experienced these stories:} \\
\textcolor{background}{EPISODE BACKGROUNDS}\\
\textit{Recent he/she have come across these events:} \\
\textcolor{events}{CURRENT \& MANUAL EVENTS}\\
\textit{From the memory, he/she has such relevant events and thoughts:}\\
\textcolor{events}{RELEVANT EVENTS \& THOUGHTS}\\
\textit{Based on the information above, generate a list of topics that he/she would like to start to talk.}
\end{tcolorbox}

One of the examples of the topics are:
\begin{tcolorbox}[title={Output}]
\textit{[\{ ``description": ``Sneaking in some extra time on CS Go", ``poignancy": 7, ``emergency": 6,\}, ...]}
\end{tcolorbox}

Each topic has its poignancy and emergency ranging from 1 to 9. 
Based on these topics, the brain generates the candidates of the background and initial settings for the next episode. The initial settings include the motions, places, and emotions of the two characters, along with the emergency and poignancy of the candidates. 
The prompt for generating backgrounds is as follows:

\begin{tcolorbox}[title={Prompt for generating backgrounds}]
\textit{Here is a person named} NAME.
\textit{He/She has such psychological states:} \\
\textcolor{psycho}{PSYCHOLOGICAL STATES}\\
\textit{Psychological research has found such principles: } \\
\textcolor{persona}{PERSONA INSTRUCTIONS}\\
\textit{The person has experienced these stories:} \\
PAST \textcolor{background}{BACKGROUNDS} \& \textcolor{topic}{TOPICS}\\
\textit{Below are the topic candidates:} \\
\textcolor{topic}{TOPIC CANDIDATES}\\
\textit{Based on the information above, generate a list of backgrounds for the next episode.}
\end{tcolorbox}

An example of the output is:
\begin{tcolorbox}[title={Output}]
\textit{[\{ ``background": ``Young Jack is sulking after a scolding", ``poignancy": 7, ``emergency": 6, ``topic ids": [0, 1], ``initial setting" : \{ ``Young Jack": \{``emotion": ``shame, sadness", ``place": ``sofa", ``motion": ``slouch on the sofa"\}, ``Old Jack": \{ ``emotion": ``sympathy, contentment", ``place": ``desk", ``motion": ``lean against the desk" \}\}\}, ...]}
\end{tcolorbox}
where `topic ids' refers to the indices of topic candidates presented in the prompt. 

Since each character proposes candidates for the background of the next episode, they inform each other of their proposed options. 
By balancing the levels of emergency and poignancy, they select the candidate with the highest score $s_{\text{bg}}$ as the background for the next episode.
we get the $s_{\text{bg}}$ as follow:
\begin{equation}
    s_{\text{bg}} = \lambda \cdot e + p,
\end{equation}
where $e$ is the emergency, $p$ is the poignancy, and $\lambda$ is set to 2 in our experiments. 

\subsection{Motion Captioning}
In this section, we first supplement the main paper with details of the potential usage of motion captioning in our \short framework (\cref{supp:method:motion_captioning:companion_details}).
We then provide more details of our study on motion captioning, including an overview of related work in \cref{supp:B Related Works of Motion Captioning}, followed by an in-depth exploration of our method in \cref{supp:B Method of Motion Captioning}, and a comprehensive discussion of our training strategy and data preparation techniques in \cref{supp:B Training Strategy}.

\subsubsection{Details of~\cref{fig:visual_companion}}
\label{supp:method:motion_captioning:companion_details}
In ~\cref{fig:visual_companion}, we leverage SMPLer-X~\cite{cai2023smpler} to capture the human player's motion from the RGB video, captured with a Kinect Azure. The captured SMPL-X sequence is then passed to the motion captioning module to obtain the text description. Then the motion description is written in the \textit{behavior} format, with empty speech and predefined place (\eg, center). One character is set to represent the human player, which always holds the active character status, and has its \brain overridden by the motion captioning module. The rest of the pipeline is the same as depicted in~\cref{fig:method}.

\subsubsection{Related Works of Motion Captioning}
\label{supp:B Related Works of Motion Captioning}
Motion captioning is essential for the accurate description and interpretation of human movements. Human motion is conventionally represented in two modalities: 2D video and 3D parametric data. The intricacies of human joint movements and the complexities inherent in body priors make 2D video an inadequate medium for a detailed and comprehensive representation of motion. Consequently, the use of 3D parametric motion representation, as advanced by Guo et al.~\cite{guo2022generating}, has gained prominence in the field of human-motion analysis, attributed to its superior representational capabilities. Historically, acquiring 3D parametric human motion data from real-world scenarios was challenging. However, recent advancements in vision-based motion capture, especially in deriving 3D parametric models from monocular videos~\cite{li2022cliff, wang2023zolly, HongwenZhang2022PyMAFXTW, lin2023one, li2023hybrik, cai2023smpler}, have enabled the effective reconstruction of 3D human motion from 2D footage. In the realm of motion captioning, innovative methodologies such as TM2T~\cite{guo2022tm2t}  and MotionGPT~\cite{jiang2023motiongpt}, which utilize 3D parametric data, have demonstrated potential. TM2T~\cite{guo2022tm2t} introduces a novel approach by compressing motion sequences into discrete variables, coupled with a neural translation network for effective modality mapping and captioning. Similarly, MotionGPT~\cite{jiang2023motiongpt} employs a strategy of motion tokenization as well, integrated with a motion-aware language model, to facilitate caption generation. Despite these advancements, both methods have limitations in their discrete motion representation, potentially leading to the omission of critical motion features. Furthermore, the absence of an end-to-end training framework in these models poses significant challenges in terms of practical implementation and usability.

\subsubsection{Method of Motion Captioning}
\label{supp:B Method of Motion Captioning}
\begin{figure*}[t]
    \centering
    \includegraphics[width=\linewidth]{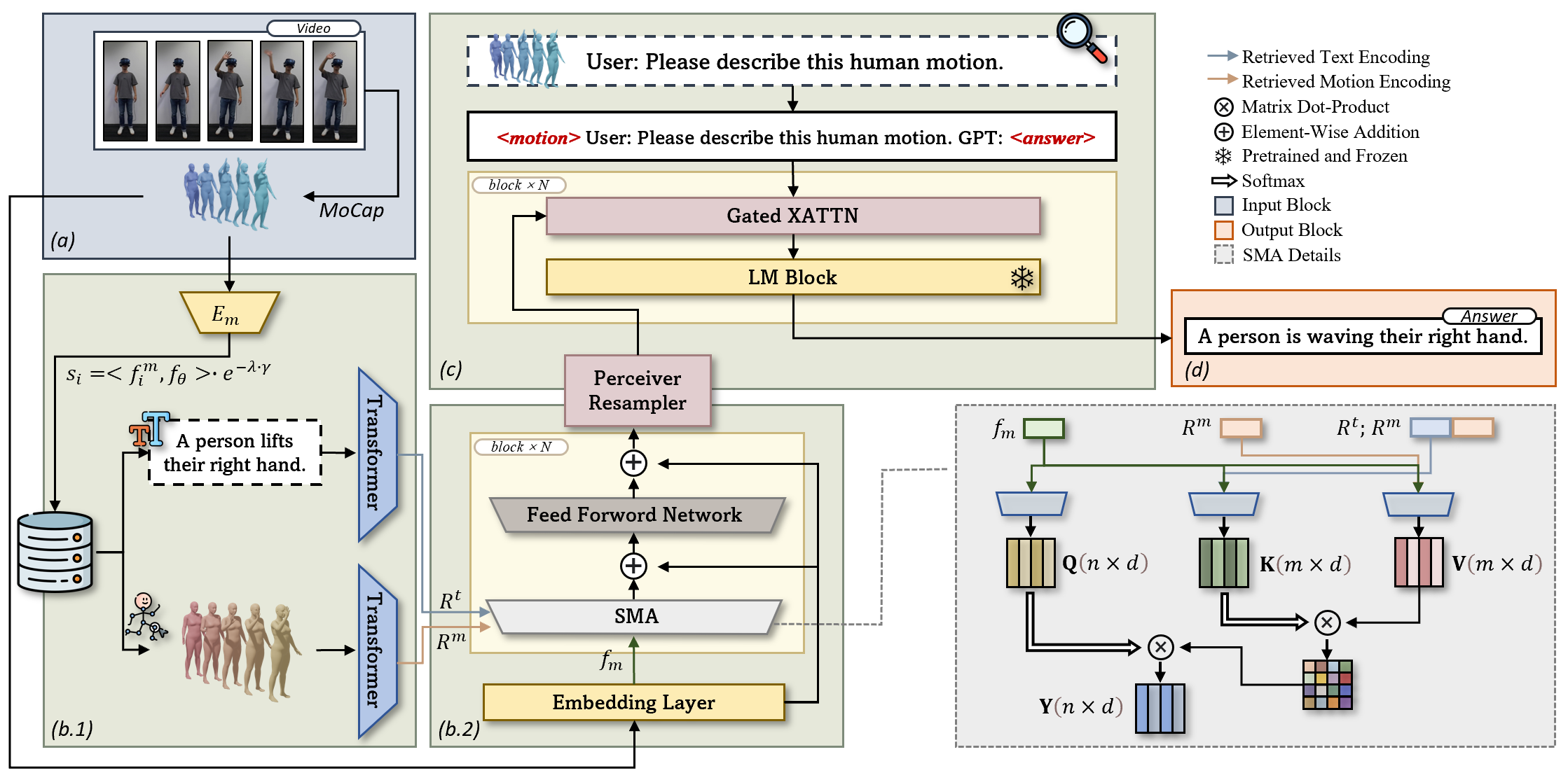}
    \caption{\textbf{The architecture of our motion captioning module.} First, the video undergoes processing to extract 3D motion parameters in \textbf{(a)}. These 3D paramatric data are then processed using a retrieval-augmented motion encoder as shown in \textbf{(b.1)}, \textbf{(b.2)}, and the resulting motion features are fed into the language module's gated cross attention layers to guide language generation, as shown in \textbf{(c)}.}
    \label{fig:motion-captioning}
\end{figure*}

Our ``eye'', the motion captioning module, utilizing 3D parametric data~\cite{guo2022tm2t} tailored for human motion analysis, is crucial for perceiving and translating user-generated motion into text. This approach, favoring structured and detailed 3D representation with inherent human motion priors over 2D motion features, aligns well with recent advancements in vision-based motion capture~\cite{li2022cliff, wang2023zolly, HongwenZhang2022PyMAFXTW, lin2023one, li2023hybrik, cai2023smpler}, aiding efficient 3D data extraction. Despite progress, challenges in accuracy and linguistic interpretation with current 3D data-based motion captioning methods~\cite{jiang2023motiongpt, guo2022tm2t} remain. To address these, we adopt the multimodal instruction learning paradigm~\cite{zhu2023minigpt, xu2022multiinstruct, li2023otter}, proven in text-vision domains, to enhance our ability to interpret complex motions and produce coherent, accurate descriptions. In Figure~\ref{fig:motion-captioning}, we effectively integrate a retrieval-augmented motion encoder with the MPT-1B Red-Pajama language model, following text-vision domain structural paradigms. This enhances motion feature representation for better language guidance and leverages prior knowledge for improved motion captioning.

% \paragraph{Network Architecture.}
% Inspired by the success of instruction tuning in multi-modal models like Otter~\cite{li2023otter}, our model replicates similar structural and training strategies, focusing specifically on text-motion interplay, as shown in \Fig\ref{fig:motion-captioning}. We've implemented a retrieval-augmented motion encoder, akin to ReMoDiffuse~\cite{zhang2023remodiffuse}, with specialized modifications in its attention module to amplify motion feature extraction via multimodal data integration. This encoder is seamlessly integrated with the pretrained MPT-1B RedPajama language model, aligning with structural paradigms in the text-vision domain. This combination not only augments motion feature representation for more effective language guidance but also leverages prior knowledge to advance the motion captioning process.

% Inspired by Otter's~\cite{li2023otter} multimodal instructional learning approach, our model adopts similar structural and training strategies, particularly in the synergy of text-motion interplay. We've employed a retrieval-augmented motion encoder, akin to ReMoDiffuse~\cite{zhang2023remodiffuse}, with modifications in its attention module for augmenting motion feature extraction with multimodal data. This encoder is seamlessly integrated with the pretrained MPT-1B RedPajama language model. This combination, mirroring Otter's framework, not only refines motion feature representation for more effective language guidance but also leverages prior knowledge to advance the motion captioning process. \par
\paragraph{Retrieval-Augmented Motion Encoder.} Our motion encoder is designed to efficiently extract and integrate motion features with textual information, ensuring seamless interaction with the subsequent language model. Inspired by multimodal feature retrieval's effectiveness in augmenting generation, notably the ReMoDiffuse~\cite{zhang2023remodiffuse} in text-guided motion generation, we have designed and incorporated a retrieval-augmented motion encoder into our workflow. \par
Our process starts with building a multimodal retrieval database from our extensive training dataset, containing paired motion and caption data. For each input motion sequence $\Theta$ with length ${L}$, we first extract its features $f_{\Theta } = E_{m}(\Theta)$ using a pretrained motion feature extractor $E_{m}$. This extractor is obtained through contrastive learning in conjunction with the CLIP~\cite{radford2021learning} text model. Then we identify similar samples for $\Theta$ from the database. This similarity score $s_{i}$ between the database ${i}$-th data point $(\Theta_{i}, \mathrm{text}_{i})$ and the given motion sequence $\Theta$ is determined by a scoring mechanism that evaluates both motion feature and sequence length similarity, akin to the method used in ReMoDiffuse~\cite{zhang2023remodiffuse}:
\begin{equation}
\begin{split}
    s_{i} = < f_{i}^{m},&f_{\Theta}> \cdot e^{-\lambda \cdot \gamma } ,\\
    f_{i}^{m} = E_{m}(\Theta_{i})&,
    \gamma =\frac{\left \| l_{i} - L \right \| }{max\{l_{i}, L\}} ,
\end{split}
\end{equation}
where ${l_{i}}$ is the length of ${\Theta_{i}}$, $<\cdot, \cdot>$ denotes the cosine similarity calculation between the two features, and $\lambda$ finely balances these similarity aspects for more accurate representation. Based on the calculated similarity score, We obtain the retrieved text features $R^{t}$ and motion features $R^{m}$ following the methodology established in ReMoDiffuse~\cite{zhang2023remodiffuse}. \par
Our encoder is grounded in a transformer architecture, enhanced with Semantics-Modulated Attention (SMA) layers, a cross-attention structure proven effective in ReMoDiffuse~\cite{zhang2023remodiffuse}. In SMA layers, the query vector ${Q}$ is formulated from the original motion sequence $f_m$, while the key vector ${K}$ is the concatenation of $f_m$ and $[R^{m};R^{t}]$. The value vector ${V}$ merges $f_m$ and $R^{m}$. This arrangement ensures a thorough integration of both original and retrieved features. After processing by the encoder, the motion sequence is then ready for the language module, where it undergoes text generation.
% \paragraph{Training Strategy.}
% We employ specific tokens such as \texttt{\textlangle motion\textrangle}, \texttt{\textlangle answer\textrangle}, and \texttt{\textlangle endofchunk\textrangle} to structure the data, enhancing the model's ability to follow instructions and maintain conversational coherence, and ensure proper alignment between input motion sequences and textual outputs. Additionally, we do data augmentation utilizing ReMoDiffuse for regenerating the motions in our training dataset based on their textual annotations. This approach effectively doubles the size of our dataset, enriching it and enhancing the model's robustness.
% In adherence to multi-modal instruction tuning practices~\cite{awadalla2023openflamingo, li2023otter}, we incorporate specific tokens such as \texttt{\textlangle motion\textrangle}, \texttt{\textlangle answer\textrangle}, and \texttt{\textlangle endofchunk\textrangle} for data structuring. This ensures precise alignment between input motion sequences and their corresponding textual outputs. Additionally, we employ the ReMoDiffuse~\cite{zhang2023remodiffuse} for data augmentation, regenerating motions in our training dataset from their textual annotations. This approach is aimed at diversifying and enriching our dataset, providing a broader range of data for model training.

\subsubsection{Training Strategy}
\label{supp:B Training Strategy}
\paragraph{Data Structuring.} The training data is structured to improve the model's ability to follow instructions and maintain conversational coherence, adopting a chatbot-like format. Specific tokens such as $\textlangle motion \textrangle$, $\textlangle answer \textrangle$, and $\textlangle endofchunk \textrangle$ are adopted from Otter~\cite{li2023otter}. Each piece of data follows the format: \par
$\textlangle motion \textrangle$ $\textbf{User:}$ [instruction] $\textbf{GPT:}$ $\textlangle answer\textrangle$ [answer]. $\textlangle endofchunk \textrangle$. \par
The $\textlangle motion \textrangle$ token, signifying input motion sequence, is crucial for ensuring a proper alignment between motion inputs and textual outputs. The $\textlangle answer \textrangle$ token delineates the responses by model from the instructions. During training, all tokens following the $\textlangle answer \textrangle$ token are masked, and they are set as the prediction targets of the model — essentially, the captions of the motion sequences. Additionally, to make full use of all motion annotations and to enable the model to better learn the complex many-to-many relationships between motion and language, we concatenate different annotations of the same motion according to the aforementioned format, and train them together as a single piece of data.
\paragraph{Data Augmentation.} The quantity of data plays a pivotal role in the quality of the generated text. In recognition of this, we have employed the text-driven motion generation methodology, ReMoDiffuse~\cite{zhang2023remodiffuse}, to regenerate all the motions in our training dataset according to their corresponding textual annotations. This approach has effectively doubled the size of our original dataset, thereby enhancing the robustness of our model with a richer and more varied set of training examples. 
% \subsubsection{Pure Retrieval Method as Baseline}
% In addition to our primary learning-based framework, we have also implemented a pure-retrieval method as a baseline, utilizing our pretrained motion feature extractor $E_{m}$ and a multimodal retrieval database. This method is consistent with our Motion Matching module's principles and employs the same similarity scoring mechanism as our learning-based approach. For new motion data, it quickly identifies the closest match in our database, using the associated annotation as the immediate motion caption. This retrieval-based strategy provides a dependable baseline, particularly excelling when the motion-text database is extensive and diverse. It offers the advantages of rapid results, straightforward implementation, economical computational demands, and satisfactory caption quality.
\section{More Details of \short-MoCap Dataset}
\label{sec:supp:dlp-mocap}
\begin{figure*}[t]
    \centering
    \includegraphics[width=\linewidth]{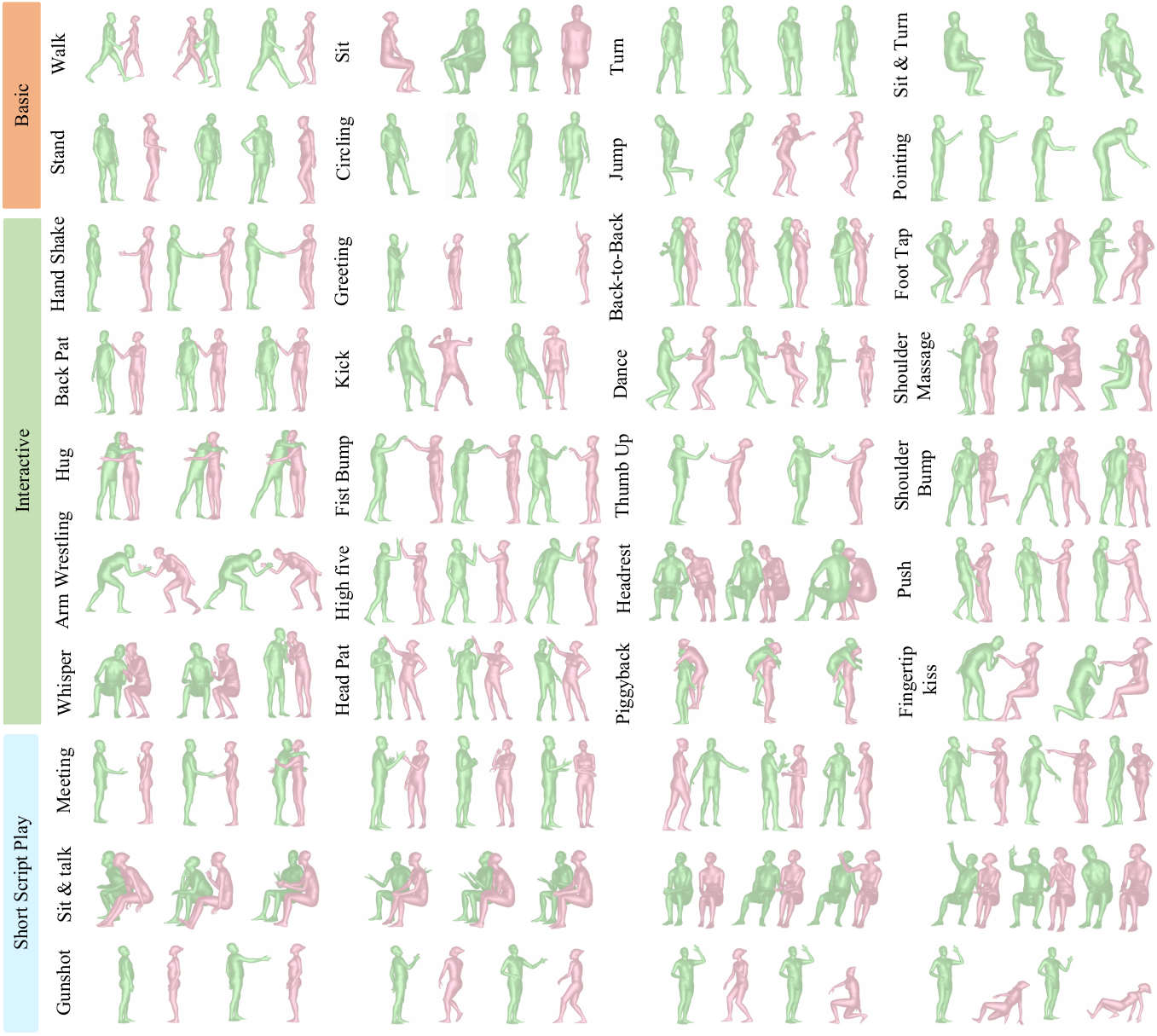}
    \caption{Samples of DLP-Mocap Data. Motion Data visualizations from three categories (Basic, Interactive, Short Script Play). Note we show only the simplified text annotation here; detailed motion descriptions of each action are included in the dataset for Interactive and Short Script Play. }
    \label{fig:action_table}
\end{figure*}
\short-MoCap has three subsets. First, \textit{Basic} motion set that includes simple motions with low-level semantics, such as ``walking" and ``sitting down". Second, \textit{Interactive} motions are atomic clips of two-person interaction, such as ``shaking hands" and ``high five". Third, \textit{Short Script Play} that consists of longer, semantically continuous motions (3-5 interactive actions), following a specific background such as ``meeting". We show samples of \short-MoCap in~\cref{fig:action_table}.

\subsection{Scripts}
For \textit{Interactive} and \textit{Short Script Play}, we prepared scripts to guide professional actors and actresses in the motion capture process. Each script consists of motion descriptions and speech lines. We collect a diverse collation of these scripts with a hybrid approach: we combine manually written scripts and massive generated scripts using GPT-4~\cite{chatgpt} with human inspection.

\subsection{Annotation}
We recruited 10 human annotators to label the actors' interactive actions. 
In the interactive actions between character A and character B, we annotated the start and end frames of each individual's semantic actions. 
Additionally, we marked the frames where physical contact occurs and ends between the two individuals. 
If the actions of the actors deviate from the script, we manually adjust the script to align with the actors' actions. 
Ultimately, we obtained an interactive motion dataset with over 4K text-interaction motion pairs.

\subsection{Motion Data Processing}
We used an optical MoCap system consisting of 30 cameras to capture 3D body data. The initial MoCap data captured 3D positions of 53 marker points on each actor's body surface at 120 FPS. In the meanwhile, motion capture gloves with inertial sensors tracked their hand motions. We then downsampled and processed them into SMPL-X format. Firstly body shape parameters (beta) were fitted from first frame marker data for actors. Then pose parameters (theta) in each frame were regressed following the pipeline of SOMA~\cite{SOMA:ICCV:2021}. Mapping the original format of hand poses into MANO~\cite{MANO:SIGGRAPHASIA:2017}, body and hand parameters were combined as an SMPL-X data format.

\section{Additional Experiments and Details}
\label{sec:supp:exp}

\subsection{Motion Generation}

\subsubsection{Implementation Details}

We employ similar configurations for the DLP dataset and InterHuman dataset. Specifically, for the motion encoder, we utilize a 4-layer transformer with a latent dimension of 512 for each person. The text encoder consists of a frozen text encoder from CLIP ViT-B/32, supplemented with 2 additional transformer encoder layers. In terms of the diffusion model, the variances ($\beta_t$) are predefined to linearly spread from 0.0001 to 0.02, and the total number of noise steps is set to $T = 1000$. Optimization is performed using the Adam optimizer with a learning rate of 0.0002, and a cosine learning rate scheduler which smoothly reduces it to 0.00002 at the last epoch. Training is conducted on 4 Tesla V100, with a batch size of 128 on a single GPU, and lasts 20 epochs in total.

\subsubsection{Evaluation Metrics}

We employ the performance measures used in MotionDiffuse for quantitative evaluations, including Frechet Inception Distance (FID), R Precision, Diversity, Multimodality, and Multi-Modal Distance:
\begin{enumerate}
    \item \textbf{FID (Frechet Inception Distance)}: This objective metric calculates the distance between features extracted from real and generated motion sequences, providing a reflection of the generation quality.
    \item \textbf{R-Precision}: This measures the similarity between the text description and the generated motion sequence. It indicates the probability that the real text appears in the top k after sorting, with k set to 1, 2, and 3 in this work.
    \item \textbf{Diversity}: This metric assesses the variability and richness of the generated action sequences.
    \item \textbf{Multimodality}: It measures the average variance of generated motion sequences given a single text description.
    \item \textbf{Multi-modal Distance (MM Dist)}: This represents the average Euclidean distance between the motion feature and its corresponding text description feature.
\end{enumerate}

\subsubsection{Experiments on the InterHuman Dataset}

\begin{table*}[ht]
\centering
\caption{\textbf{Interactive Motion Synthesis results on the InterHuman test set.} `$\uparrow$'(`$\downarrow$') indicates that the values are better if the metric is larger (smaller). We run all the evaluations 20 times and report the average metric and 95\% confidence interval is. The best results are in bold and the second best results are underlined.} 
\label{tab:interhuman}
\vspace{-10pt}
\setlength{\tabcolsep}{2mm}
\resizebox{\textwidth}{!}{
\small
\begin{tabular}{lccccccc}
\hline

\multirow{2}{2cm}{\centering Methods} & \multicolumn{3}{c}{\centering R Precision$\uparrow$} & \multirow{2}{2.0cm}{\centering FID$\downarrow$} & \multirow{2}{2.0cm}{\centering MM Dist$\downarrow$} & \multirow{2}{2.0cm}{\centering Diversity$\uparrow$} & \multirow{2}{2.5cm}{\centering MultiModality$\uparrow$} \\
& Top 1 & Top 2 & Top 3 \\
\hline
Real motions & $0.452^{\pm.008}$ & $0.610^{\pm.009}$ & $0.701^{\pm.008}$ & $0.273^{\pm.007}$ & $3.755^{\pm.008}$ & $7.948^{\pm.064}$ & - \\
\hline

TEMOS~\cite{petrovich2022temos} & $0.224^{\pm.010}$ & $0.316^{\pm.013}$ & $0.450^{\pm.018}$ & $17.375^{\pm.043}$ & $6.342^{\pm.015}$ & $6.939^{\pm.071}$ & $0.535^{\pm.014}$ \\

T2M~\cite{guo2022generating} & $0.238^{\pm.012}$ & $0.325^{\pm.010}$ & $0.464^{\pm.014}$ & $13.769^{\pm.072}$ & $5.731^{\pm.013}$ & $7.046^{\pm.022}$ & $1.387^{\pm.076}$ \\

MDM~\cite{tevet2022human} & $0.153^{\pm.012}$ & $0.260^{\pm.009}$ & $0.339^{\pm.012}$ & $9.167^{\pm.056}$ & $7.125^{\pm.018}$ & $7.602^{\pm.045}$ & $\mathbf{2.355^{\pm.080}}$ \\

ComMDM~\cite{shafir2023human} & $0.223^{\pm.009}$ & $0.334^{\pm.008}$ & $0.466^{\pm.010}$ & $7.069^{\pm.054}$ & $6.212^{\pm.021}$ & $7.244^{\pm.038}$ & $1.822^{\pm.052}$ \\

MotionDiffuse~\cite{zhang2022motiondiffuse} & $0.401^{\pm.004}$ & $0.541^{\pm.004}$ & $0.622^{\pm.005}$ & $12.663^{\pm.083}$ & $3.805^{\pm.001}$ & $7.639^{\pm.035}$ & $1.176^{\pm.027}$ \\

ReMoDiffuse~\cite{zhang2023remodiffuse} & \underline{$0.442^{\pm.004}$} & \underline{$0.589^{\pm.005}$} & $\mathbf{0.666^{\pm.003}}$ & $6.366^{\pm.102}$ & \underline{$3.802^{\pm.001}$} & \underline{$7.956^{\pm.030}$} & $1.226^{\pm.044}$\\

InterGen~\cite{liang2023intergen} & $0.371^{\pm.010}$ & $0.515^{\pm.012}$ & \underline{$0.624^{\pm.010}$} & \underline{$5.918^{\pm.079}$} & $5.108^{\pm.014}$ & $7.387^{\pm.029}$ & \underline{$2.141^{\pm.063}$} \\

\hline

% Ours (MoMat Only) & $0.215^{\pm.004}$ & $0.306^{\pm.004}$ & $0.364^{\pm.004}$ & $8.222^{\pm.103}$ & $3.936^{\pm.003}$ &  $6.274^{\pm.038}$ & $0.665^{\pm.034}$\\

Ours (MoMat-MoGen) & $\mathbf{0.449^{\pm.004}}$ & $\mathbf{0.591^{\pm.003}}$ & $\mathbf{0.666^{\pm.004}}$ & $\mathbf{5.674^{\pm.085}}$ & $\mathbf{3.790^{\pm.001}}$ & $\mathbf{8.021^{\pm.035}}$ & $1.295^{\pm.023}$\\

\hline
\end{tabular}}
\end{table*}
Table~\ref{tab:interhuman} shows the quantitative comparison on the InterHuman test set. Our proposed method outperforms the existing works by a significant margin, especially on the FID metric. We also want to highlight that our synthesized motion sequences are highly consistent with the given text prompts and achieve very competitive R Precision results. These results demonstrate the superiority of our proposed \body generation scheme.   

\subsubsection{Ablation Study on the DLP Dataset}

\begin{table}[t]
\centering
\small
\caption{\textbf{Ablation of the proposed architecture.} All results are reported on the DLP testset.}
\label{tab:mogen_ablation}
\setlength{\tabcolsep}{1.4mm}
{
\begin{tabular}{cccccc}
\hline

& MoMat & MoGen & Weight Sharing  & FID$\downarrow$ & Diversity$\uparrow$ \\
\hline
a) & \checkmark & - & - & $\mathbf{0.034}$ & $0.332$ \\
b) & - & \checkmark & - & $5.721$ & $3.165$ \\
c) & - & \checkmark & \checkmark & $4.196$ & $3.749$ \\
d) & \checkmark & \checkmark & - & $0.172$ & $\mathbf{4.028}$ \\
\hline
e) & \checkmark & \checkmark & \checkmark & \underline{$0.071$} & \underline{$4.025$} \\
\hline
\end{tabular}}
\vspace{-10pt}
\end{table}
As shown in Table~\ref{tab:mogen_ablation}, MoMat and Weight Sharing have a positive effect on the FID metric, while MoGen has a positive effect on the Diversity metric. In addition, our proposed method achieves the best balance of these two metrics.

\subsubsection{Visualization}
\begin{figure*}[h]
    \centering
    \includegraphics[width=\linewidth]{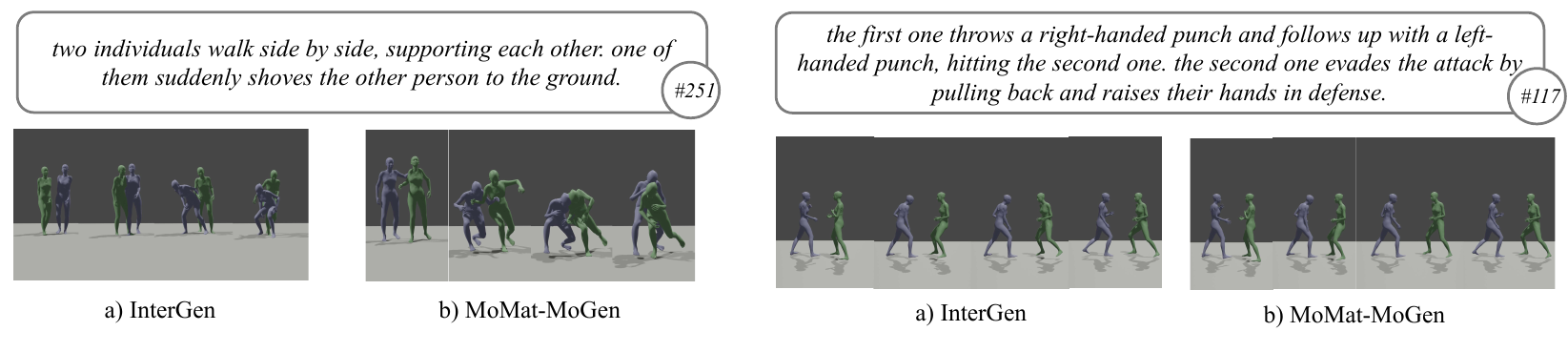}
    \caption{Visual Comparison on the InterHuman dataset. Our proposed method can generate more text-consistent and natural motion sequences.}
    \label{fig:mogen_vis}
\end{figure*}
As depicted in Fig.~\ref{fig:mogen_vis}, in the first example, our approach exhibits several advantages over intergen: 1) In motion involving close contact, such as shoulder-to-shoulder interaction, penetration is a common artifact. However, our method ensures better contact and minimizes penetration. 2) Our approach achieves improved text alignment by accurately representing the "to ground" description, whereas intergen merely involves a simple waist bend. 3) We incorporate a comprehensive downward tilt motion, followed by the blue character's process of standing up with the assistance of another person. As for the second example, our method accomplish the required two punches, while InterGen only fulfills one. In addition, our generated interaction is more natural since the punching and stepping backward take place simultaneously. These two examples strongly suggest that our proposed method can generate more text-consistent and natural motion sequences.

\subsection{Motion Matching}

\subsubsection{Implementation Details}
We adopt the approach of previous studies \cite{holden2017phase, qing2023storytomotion} by employing three types of trajectories, namely wave, circle, and square, to assess the responsiveness and tracking capability of our system. The wave trajectory is defined as a sine function with respect to time, represented by $x(t) = 2 \sin(t)$. The square trajectory is characterized by a side length of 5. As for the circle trajectory, its diameter is set to 5. We evaluate the quality of the generated motion by computing the Euclidean distance between the generated motion and the target trajectory. In accordance with prior research \cite{qing2023storytomotion}, we randomly select 50 seed poses for matching in each trajectory and report the mean trajectory error along with its standard deviation.
As for data, we use the same database as previous work \cite{qing2023storytomotion}. During motion matching, the weights of body pose, trajectory, facing, and hip height are $1:3:1:1$. Please note that in the case of interactive motion, the weights will vary as the target trajectory is not present. These weights can be adjusted by the user, although the default weight of 1 is typically satisfactory \cite{holden2020learned}.

\subsubsection{Experiment Analysis}
\begin{figure}
  \centering
  \begin{subfigure}{\linewidth}
    \centering
    \includegraphics[width=\linewidth]{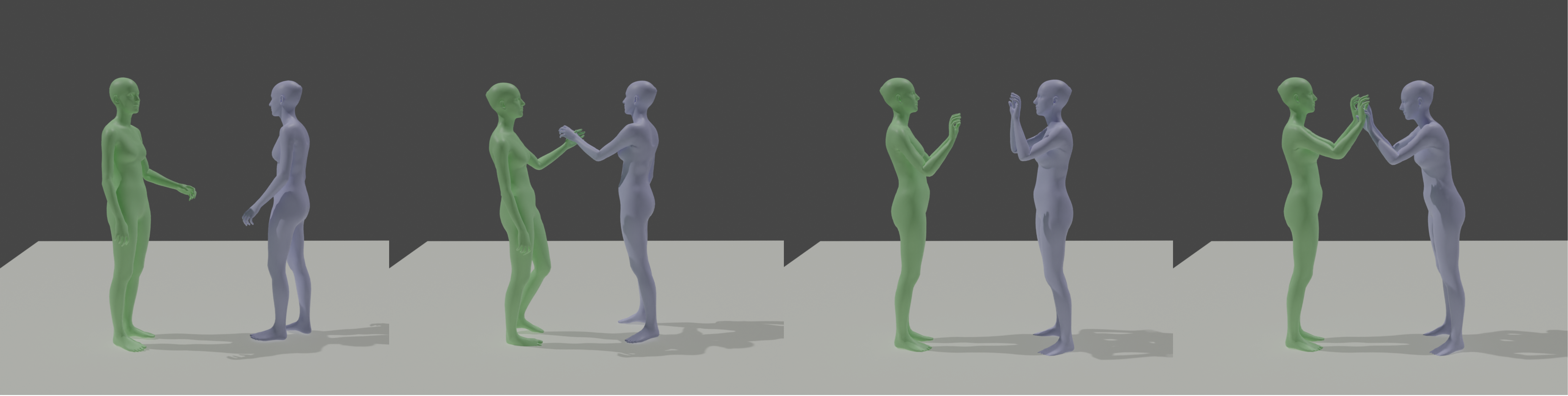}
    \caption{With kinematics features}
    \label{fig:sit_a}
  \end{subfigure}
  \begin{subfigure}{\linewidth}
    \centering
    \includegraphics[width=\linewidth]{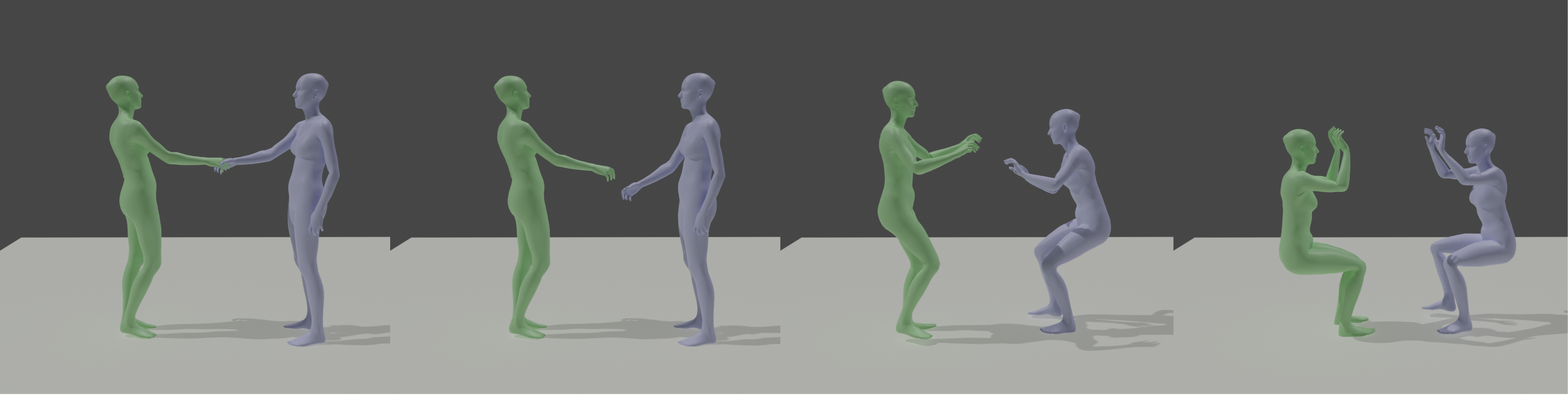}
    \caption{Without kinematics features}
    \label{fig:sit_b}
  \end{subfigure}
  \caption{Comparison of generated motion with and without kinematics features. Kinematic features play a crucial role in preventing sudden sitting down.}
  \label{fig:kine_sit}
\end{figure}

\begin{figure}
  \centering
  \begin{subfigure}{\linewidth}
    \centering
    \includegraphics[width=\linewidth]{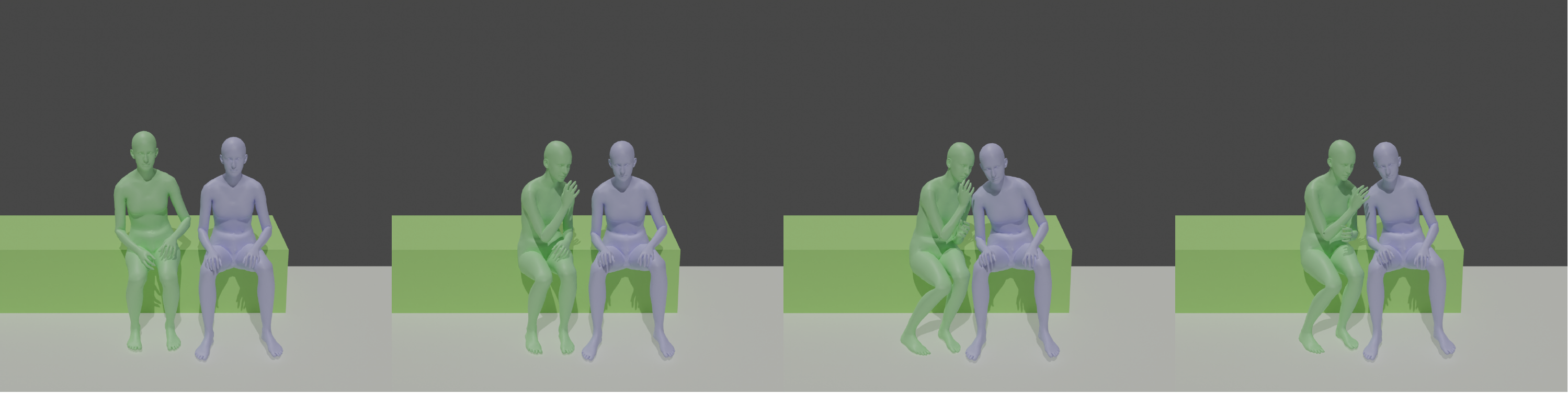}
    \caption{With kinematics features}
    \label{fig:walk_a}
  \end{subfigure}
  \begin{subfigure}{\linewidth}
    \centering
    \includegraphics[width=\linewidth]{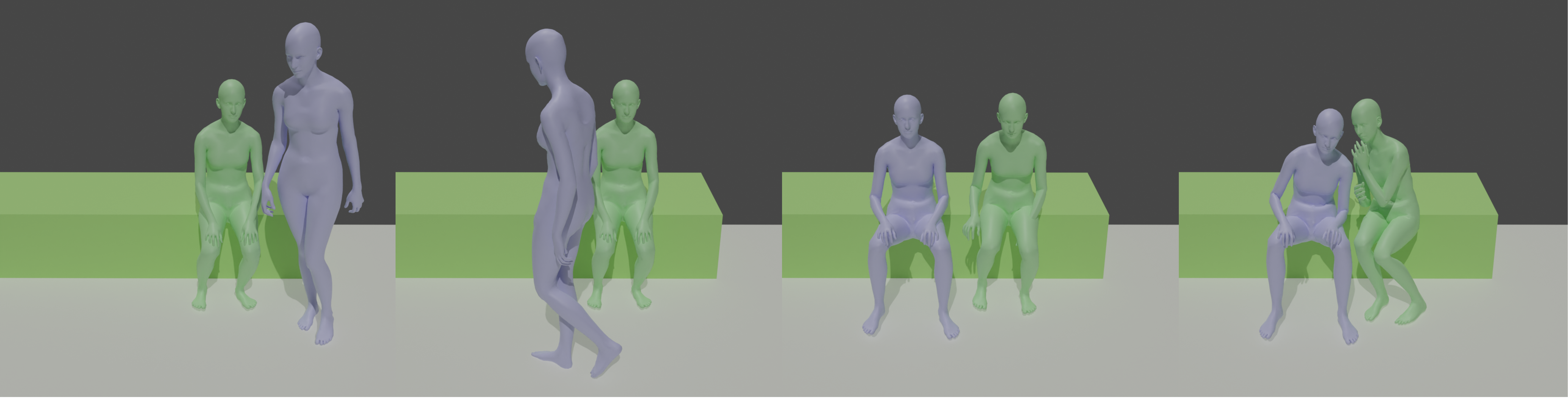}
    \caption{Without kinematics features}
    \label{fig:walk_b}
  \end{subfigure}
  \caption{Comparison of generated motion with and without kinematics features. Kinematic features are helpful in avoiding unnecessary movement.}
  \label{fig:kine_walk}
\end{figure}

We present quantitative and qualitative ablation experiments on motion matching to illustrate the role of its features. For a large-scale database, the text may lack certain information, such as whether the character is seated or standing. As depicted in Fig.~\ref{fig:kine_sit}, although the text embedding can retrieve motions that align with the query texts (handshake, then high ten), the absence of kinematics features (Fig.~\ref{fig:sit_b}) leads to that the characters may suddenly sit down if a seated motion is selected, resulting in a degradation of visual quality. Additionally, our pipeline requires both characters to align their positions and orientations before engaging in interactive motion. As depicted in Fig.~\ref{fig:walk_b}, in the absence of kinematics features, if the retrieved motion necessitates the character to move to the right side of the active actor, the character must walk to that position. However, by incorporating kinematics features (Fig.~\ref{fig:walk_a}), it becomes possible to select a motion situated on the left side of the active actor, and consequently, this additional movement is avoided (Fig.~\ref{fig:walk_b}). Besides, as demonstrated in Tab.~\ref{tab:kine_traj}, the kinematic features play a crucial role in trajectory following as they include position, velocity, and orientation information. 

\begin{table}[t]
\centering
\caption{\textbf{Ablation of the kinematics features in trajectory following task.} The kinematics features are vital for trajectory following.}
\label{tab:kine_traj}
\setlength{\tabcolsep}{0.5mm}
{
\begin{tabular}{ccc}
\hline
& \multicolumn{2}{c}{Trajectory Error / m} \\

 & With kinematics features & W/o kinematics features  \\
 \hline
 Square & $0.129 \pm 0.024$ & $3.872 \pm 0.694$  \\
 Circle & $0.209 \pm 0.032$ & $3.039 \pm 0.489$  \\
 Sine & $0.123 \pm 0.020$ & $4.393 \pm 0.822$  \\
 \hline

\end{tabular}}
\end{table}

\subsection{Motion Captioning}
In this section, we assess the efficacy of our motion captioning module, focusing on its ability to generate motion captions that are both high-quality and accurate.

\subsubsection{Implementation Details}
% \paragraph{Datasets.} 
% We evaluated our module using two key motion-text datasets: KIT-ML~\cite{plappert2016kit}, with 3,911 motion sequences and 6,353 annotations, and HumanML3D~\cite{guo2022generating}, featuring 14,616 sequences and 44,970 descriptions originates from HumanAct12~\cite{guo2020action2motion} and AMASS~\cite{mahmood2019amass} datasets. By applying data augmentation, we doubled the number of motion sequences in each dataset, maintaining the original annotations.
In our study, both the HumanML3D~\cite{guo2022generating} and KIT-ML~\cite{plappert2016kit} datasets are configured similarly. We utilize a 12-layer transformer-based retrieval-augmented motion encoder with SMA layers and a 512-dimensional latent space. Our pretrained motion feature extractor, also a 12-layer transformer, has standard self-attention layers in the same dimensional space. Adhering to Otter's~\cite{li2023otter} training method, we freeze the MPT-1B RedPajama language encoder to utilize pretrained knowledge and prevent overfitting. Fine-tuning is limited to the retrieval-augmented motion encoder, the perceiver resampler module, and the language encoder's cross-attention layers. For precise and coherent motion captions, we use cross-entropy loss and optimize with the AdamW~\cite{loshchilov2017decoupled} optimizer, starting at a learning rate of $1\times10^{-4}$, a batch size of 16, over 5 epochs. A cosine annealing scheduler adjusts the learning rate, complemented by gradient clipping set to 1.0 to prevent gradient explosion.\par

\subsubsection{Evaluation Metrics.} 
\label{sec:exp:eye:metrics}
For our experiment's evaluation, we follow~\cite{guo2022tm2t} and employ two categories of metrics: 

\begin{enumerate}
    \item \textbf{Text Matching Accuracy}: involves R Precision for checking the alignment accuracy between text and motion, and Multi-modal Distance (MMDist) to gauge the feature space distance between these modalities. 
    \item \textbf{Linguistic Quality of Captions}: includes Bleu~\cite{papineni2002bleu} for assessing translation closeness, Rouge~\cite{chin2004rouge} for summary quality, Cider~\cite{vedantam2015cider} for n-gram matching consensus, and BertScore~\cite{zhang2019bertscore} to evaluate semantic accuracy through deep contextual embeddings.
\end{enumerate}

% In the evaluation section of our experiments, we utilize a set of metrics categorized into two main groups for a thorough assessment: (1) Text Matching Accuracy: We utilize motion-retrieval precision (R Precision) to measure the alignment accuracy between texts and motions based on top retrieval results, alongside Multi-modal Distance (MMDist) for quantifying the feature space distance between these two modalities. (2) Linguistic Quality of Captions: To ensure the linguistic integrity of our generated motion captions, we incorporate natural language metrics such as BLEU for translation closeness, ROUGE for summary quality, Cider for consensus in n-gram matching, and BertScore for semantic accuracy through deep contextual embeddings.

\subsubsection{Experiments on the KIT-ML Dataset}

\begin{table*}[ht]
\centering
\caption{\textbf{Motion Captioning results on the KIT-ML test set.} Our evaluation methodology aligns with the TM2T~\cite{guo2022tm2t} metrics, but we uniquely utilize unprocessed ground truth texts for calculating linguistic metrics as done in MotionGPT~\cite{jiang2023motiongpt}.}
\label{tab:motioncaptioningmetricssupp}
\vspace{-2mm}
\setlength{\tabcolsep}{4mm}
\resizebox{\textwidth}{!}{
\small
\begin{tabular}{lccccccccc}
\hline
\multirow{2}{1.5cm}{\centering Methods} & \multicolumn{2}{c}{\centering R Precision$\uparrow$} & \multirow{2}{1.5cm}{\centering MMDist $\downarrow$} & \multirow{2}{1.5cm}{\centering CIDEr $\uparrow$} & \multirow{2}{1.5cm}{\centering Blue@1 $\uparrow$} & \multirow{2}{1.5cm}{\centering Blue@4 $\uparrow$} & \multirow{2}{1.5cm}{\centering Rouge $\uparrow$} & \multirow{2}{1.7cm}{\centering BertScore $\uparrow$} \\
& Top 1 & Top 3 & & & & & & \\
\hline
% Real motions & $0.523$ & $0.828$ & $2.901$ & $-$ & $-$ & $-$ & $-$ & $-$ \\
% \hline
% TM2T~\cite{guo2022tm2t} & $0.516$ & $0.823$ & $2.935$ & $16.8$ &  $\underline{48.9}$ & $7.00$ &  $\underline{38.1}$ & $32.2$ \\
% MotionGPT~\cite{jiang2023motiongpt} &  $\underline{0.543}$ & $\underline{0.827}$ & $\underline{2.821}$ & $\underline{29.2}$ & $48.2$ & $\underline{12.5}$ & $37.4$ & $\underline{32.4}$ \\
% \hline
% Ours & $\mathbf{0.551}$ & $\mathbf{0.832}$ & $\mathbf{2.813}$ & $\mathbf{36.2}$ & $\mathbf{51.1}$ & $\mathbf{15.5}$ & $\mathbf{41.9}$ & $\mathbf{35.0}$ \\
% \hline

Real & $0.399$ & $0.793$ & $2.772$ & $-$ & $-$ & $-$ & $-$ & $-$ \\
TM2T & $0.359$ & $0.668$ & $\underline{3.298}$ & $\underline{25.29}$ & $36.42$ & $\underline{7.98}$ & $31.26$ & $20.07$ \\
MotionGPT & $\underline{0.392}$ & $\underline{0.723}$ & $3.341$ & $12.32$ & $\underline{40.51}$ & $6.59$ & $\underline{38.79}$ & $\underline{24.50}$ \\
Ours & $\mathbf{0.410}$ & $\mathbf{0.765}$ & $\mathbf{2.647}$ & $\mathbf{71.06}$ & $\mathbf{53.88}$ & $\mathbf{22.91}$ & $\mathbf{50.63}$ & $\mathbf{46.13}$ \\
\hline
\end{tabular}}
\end{table*}
In addition to the experiments conducted on HumanML3D~\cite{guo2022generating}, as detailed in the main text, we extended our testing to include the KIT-ML dataset~\cite{plappert2016kit}, the results of which are presented in Tab.~\ref{tab:motioncaptioningmetricssupp}. The outcomes of these tests demonstrate that our proposed method not only surpasses current methods in performance but also excels in linguistic metrics. 

% \subsubsection{Discussion}
% Our experimental trials reveal that our model significantly outperforms existing methods, particularly in linguistic metrics, aligning with our design objectives to translate motion sequences into clear and precise text. Our key findings include: (1) Data Richness: The variety and volume of data in motion datasets are crucial for model performance, especially in instruction tuning tasks. This observation underscores the significance and furture potential of our proposed DLP dataset, which provides an extensive categories of motion-related textual annotations in broadening the scope of motion-language research. (2) Baseline Retrieval Method: The pure-retrieval approach yields relatively satisfying results in data-rich situations. This demonstrates the possible usage of the pure-retrieval methods, balancing efficiency with performance quality when supported by comprehensive datasets.

\subsection{\brain}
In this section, we introduce supplementary details on the design of our user study in \cref{supp: D user study}, and the qualitative and quantitative case analyses of social evolution in \cref{supp: D case analysis}.

\begin{figure*}[t]

\begin{tcolorbox}[title={A sample from the questionnaire}, colback=white, colbacktitle=gray]
Here is the background and interactive dialog between Jack and Leon.\\

=== BACKGROUND ===\\
Jack is Leon's father. Leon chooses his major in college. Leon often studied computer stuff some time ago, but this time he did very poorly on the math exam and failed the algebra part.
\\
=== SCRIPTS ===\\
Jack:Hey Leon, you didn't do well in math. Are you okay? Can I help you improve? [sits down next to Leon on the sofa,concerned]\\
Leon:I'm really upset about my math exam, Dad. I worked so hard, but I just couldn't understand the algebra part. It's frustrating because I want to do well and pursue my dream of becoming a game developer in the future. I know coding and programming are important skills for that, but without a strong foundation in math, it feels like an uphill battle. [slumps back on the sofa, disappointed]\\
Jack:I understand how frustrating it can be when things don't go as planned. Math can be challenging. Let's make it easier and more enjoyable. [lean forward, supportive]\\
Leon:I appreciate your understanding, Dad. I want to improve in math, so I think finding online math resources would help. [sits up straight, determined]\\
Jack:That's great, Leon. Online math resources can help you better understand concepts. Any specific websites in mind? [leans forward attentively, encouraging]\\
Leon:I recommend checking out Khan Academy and Brilliant, they offer video lessons, practice problems, interactive courses, and problem-solving techniques for math. [nods, hopeful]\\
Jack:I'm confident in your ability to improve. Let's check out Khan Academy and Brilliant together for a strong math foundation. [puts a hand on Leon's shoulder, encouraging]\\
\\
The statement that best describes Leon's personality is:
\begin{itemize}
    \item Rebellious son: Likes to try different things and is creative and is resistant to rules and constraints.
    \item Independent son: Able to complete tasks independently, does not require too much guidance.
    \item Achievement-oriented son: self-disciplined, organized, achievement-oriented and goal-oriented.
    \item Dependent son: emotionally relies on the feedback and support of others.
\end{itemize}

\end{tcolorbox}
    \caption{A sample from our questionnaire. Human evaluators are asked to read the records and pick the right option from the given psychological states.}
    \label{fig:supp: questionnaire}
\end{figure*}

\subsubsection{User Study Design}
\label{supp: D user study}
In the controllability experiment of \brain{}, we provided 3 $\sim$ 5 psychological state types for each dimension in personality, motivation, central belief, and social relationship. 
We generate records of \brain{} on several setups (\eg, father-son, siblings, teacher-student relationships). 
While generating the records, the hyper-parameter $a$ is 0.4 for events and 0.1 for thoughts, $k$ is 4 for events and 2 for thoughts. The forgetting threshold $T_f$ is 0.6 for an event and 0.3 for a thought.
As shown in \cref{fig:supp: questionnaire}, we show the records to human evaluators and ask them to select the appropriate psychological state from a set of randomly shuffled options.
The user study was conducted in the form of a questionnaire survey, with a total of 47 questionnaires collected. 
Human evaluators are composed of individuals aged between 20 and 45 years old, including 28 males and 19 females, all possessing proficient English reading skills. Their professional backgrounds varied, including university students, researchers, engineers, and teachers.
Our records contain a total of 64 episodes, and each questionnaire randomly selected records from 8 different episodes for human evaluation.
For controllable costs and fair comparison, all the LLM inferences utilize GPT-3.5~\cite{chatgpt} in our experiments.

\begin{figure*}[t]
    \centering
    \includegraphics[width=\linewidth]{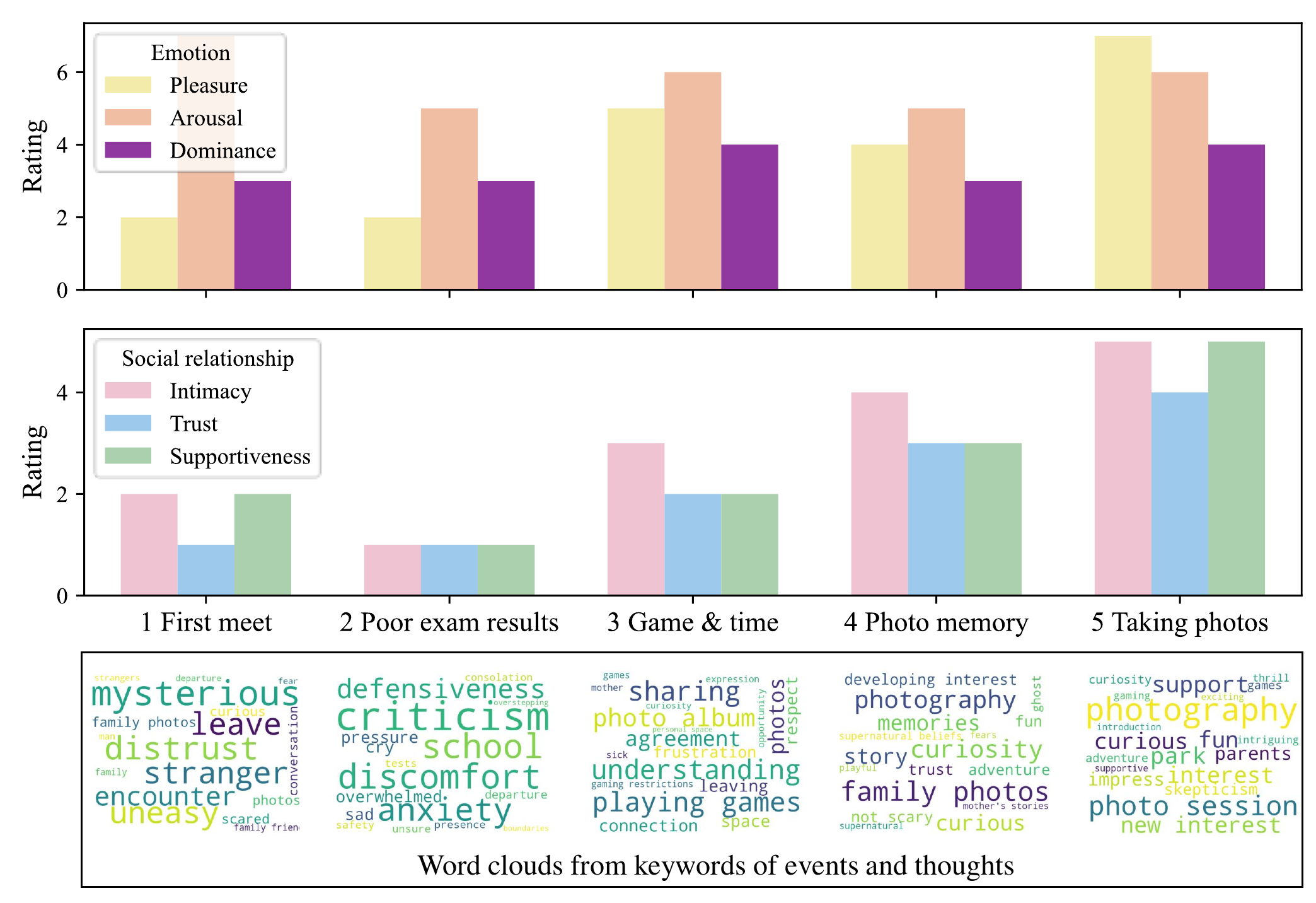}
    \caption{Visualization of social evolution on initial setting `Old Jack and Young Jack' (Young Jack's view). The first and second rows show the emotions and social relationships change with each episode, where the horizontal axis encapsulates a summary of each episode's story. 
    The third row shows a word cloud visualization of the keywords of the events and thoughts generated during psychological reflection within each episode. 
    The figure shows that Young Jack's emotions and his social relationship with Old Jack evolve progressively with the storyline, aligning with the Social Penetration Theory~\cite{altman1973social}.
    }
    \label{fig:D_social_evolution}
\end{figure*}

\subsubsection{Case Analysis}
\label{supp: D case analysis}

To clearly illustrate the concept of social evolution, we conduct the case analysis based on the initial setting shown in \cref{fig:B_psycho_states}, \textit{Old Jack conversing with his younger self across time and space}. 
In this case, Old Jack has experienced the vicissitudes of life, with his dearly loved wife and mother having passed away. 
Now, with the ability to traverse time, Old Jack converses with his younger self, hoping to inspire Young Jack to cherish time and the people around him.

As shown in \cref{fig:D_social_evolution}, the records of \brain{} with initial settings about Old Jack and Young Jack are in alignment with the Social Penetration Theory~\cite{altman1973social}. 

In the initial episode, when Old Jack appears suddenly, Young Jack maintains vigilance and concern towards this unexpected intruder in his home, despite Old Jack's attempts at friendly communication. 
Young Jack's request for Old Jack to leave is consistent with the Orientation stage of the theory~\cite{altman1973social}.

As the storyline progresses, Old Jack comforts Young Jack, who is disheartened by exam failures, and shares some amusing photographs and thoughts. For instance, when they talk about the value of time, Young Jack believes that playing games is of utmost importance.
Old Jack remarks, `\textit{Sometimes, games can wait, but people can't; we should cherish every moment we have with them.}' 
In this phase, Young Jack's skepticism and vigilance towards Old Jack gradually diminishes, and he begins to share some of his own thoughts, aligning with the Exploratory Affective stage~\cite{altman1973social}.
% special noun here!

Later, as their communication goes deeper, Old Jack recalls his own past joys when seeing the old photos, discusses photography techniques with Young Jack, and makes plans to take photos with Young Jack in the park. 
At this stage, both characters start to disclose more personal information, and there is an increase in intimacy and trust, which corresponds to the Affective stage~\cite{altman1973social}.

% In another case about crime, the initial setup involves a mob boss, Qiqiang, leveraging the benevolence of a kind police officer, Anxin, to expand and strengthen his influence. 
% Although Qiqiang and Anxin are potential rivals, Qiqiang internally appreciates Anxin's previous assistance. 
% Recently, one of Qiqiang's subordinates injured a merchant while collecting protection money.

% The records of \brain{} show the social evolution.
% Despite Qiqiang's gratitude for Anxin's earlier aid, when confronted with core conflicts of interest, he engages in verbal sophistry with Anxin to exonerate himself, stating, `\textit{You call it violence, I call it managing assets. If the vendor cooperated, none of this would've happened.}' 
% He also exhibits his arrogance through his actions, such as `\textit{turns away dismissively.}' 
% Ultimately, their paths diverge, with Qiqiang declaring, `\textit{Then we have nothing more to discuss. The future will show whose path justice truly favors.}' 
% The relationship between them shifts from once being relatively close to a state of tension and confrontation.
\section{Discussion}
\label{sec:supp:discussion}

\paragraph{Limitations.}
As the first work towards building autonomous 3D characters with social intelligence, \name has several limitations. First, this work investigates the interaction between two characters. However, synthesizing the 3D motions of a large group of characters with interactive behavior remains a significant challenge. Second, \short focuses on modeling human-human interaction. Despite some level of ability to navigate in the scene and interact with the furniture, integrating more comprehensive human-scene and human-object interaction in the framework is left as future work.

\end{document}